\documentclass{article}

\usepackage{iclr2017_arXiv,times}
\usepackage{hyperref}
\usepackage{url}

\usepackage{tabularx}
\usepackage{multirow}

\usepackage{algorithm2e}

\usepackage{graphicx} % Allows including images

\usepackage{bm}
\usepackage{amsmath}

\usepackage{xcolor}

\SetCommentSty{mycommfont}

\usepackage{hyperref}
\hypersetup{
    colorlinks,
    citecolor=black,
    filecolor=black,
    linkcolor=black,
    urlcolor=black
}

\usepackage{multirow}

\usepackage{amssymb} % for checkmark
\usepackage{rotating} % for vertical text

\usepackage{lipsum}

\usepackage[english]{babel}
\usepackage[autostyle]{csquotes}

\usepackage{makeidx}
\makeindex

\usepackage{makecell}

\title{Reinforcement Learning Applications}

\author{
  Yuxi Li (yuxili@gmail.com)\\
  }

\begin{document} 

\maketitle

\newpage
\tableofcontents
\newpage

\section{Introduction}
\label{introduction}

What is the most exciting AI news in recent years?

AlphaGo!

\vspace{3mm}

What are the underlying techniques?

Deep learning and reinforcement learning!

\vspace{3mm}

What are applications of reinforcement learning?

A lot!

\vspace{3mm}

Reinforcement learning (RL) has been making steady progress, with decades of research and development.
RL has achieved spectacular achievements, and will be achieving more and more.

%In the following, we briefly introduce RL, about successful stories, basics, parlance, an example, issues, 
%the ICML 2019 Workshop on RL for Real Life, how to use it, study material and an outlook.
%Then, we discuss the outline of the book.
%At the end of each chapter, we present an annotated bibliography to discuss relevant reference.

\subsection{Successful Stories}

We have witnessed breakthroughs, such as Atari games, AlphaGo (AlphaGo Zero, AlphaZero), and DeepStack/Libratus.
Each of them represents a big family of problems and the underlying techniques can be applied to a large number of applications. 
Atari games are for single-player games and single-agent control in general,
which ignited the current round of popularity of deep RL. 
AlphaGo is for two-player perfect information zero-sum games. 
AlphaGo made a phenomenal achievement on a very hard problem, and set a landmark in AI. 
DeepStack/Libratus is for two-player imperfect information zero-sum games, a family of problems which are inherently difficult to solve. 
DeepStack/Libratus also set a milestone in AI.

Deepmind AlphaStar defeats top human players at StarCraft II.
Deepmind has achieved human-level performance in the multi-player game Catch The Flag. 
OpenAI Five defeats good human players at Dota 2. 
OpenAI trains Dactyl for a human-like robot hand to dextrously manipulate physical objects. 
Google AI applies RL to data center cooling, a real-world physical system. 
DeepMimic simulates humanoid to perform highly dynamic and acrobatic skills. 
RL has also been applied to chemical retrosynthesis and computational de novo drug design. 
And  so  on and so forth.

We have also seen applications of RL in products and services. 
AutoML attempts to make AI easily accessible. 
Google Cloud AutoML provides services like the automation of neural architecture design, device placement and data augmentation. 
Facebook Horizon has open-sourced an RL platform for products and services like notification delivery and streaming video bit rates optimization. 
Amazon has launched a physical RL testbed AWS DeepRacer, together with Intel RL Coach.

%The techniques underlying these achievements, namely, deep learning, RL, Monte Carlo tree search (MCTS), and self-learning, will have wider and further implications and applications.

\subsection{A Brief Introduction to RL}

Machine learning is about learning from data and making predictions and/or decisions.
We usually categorize machine learning as supervised learning, unsupervised learning, and reinforcement learning. 
In supervised learning, there are labeled data; 
in unsupervised learning, data are not labeled. 
Classification and regression are two types of supervised learning problems, with categorical and numerical outputs, respectively.
In RL, there are evaluative feedbacks but no supervised labels. 
Evaluative feedbacks can not indicate whether a decision is correct or not, as labels in supervised learning.
RL has additional challenges like credit assignment, stability, and exploration, comparing with supervised learning.
Deep learning (DL), or deep neural networks (DNNs), can work with/as these and other machine learning approaches. 
Deep learning is part of machine learning, which is part of AI. 
Deep RL is an integration of deep learning and RL.
%, e.g., by approximating value function or policy with DNNs. 

\begin{figure}
\centering
\includegraphics[width=0.5\linewidth]{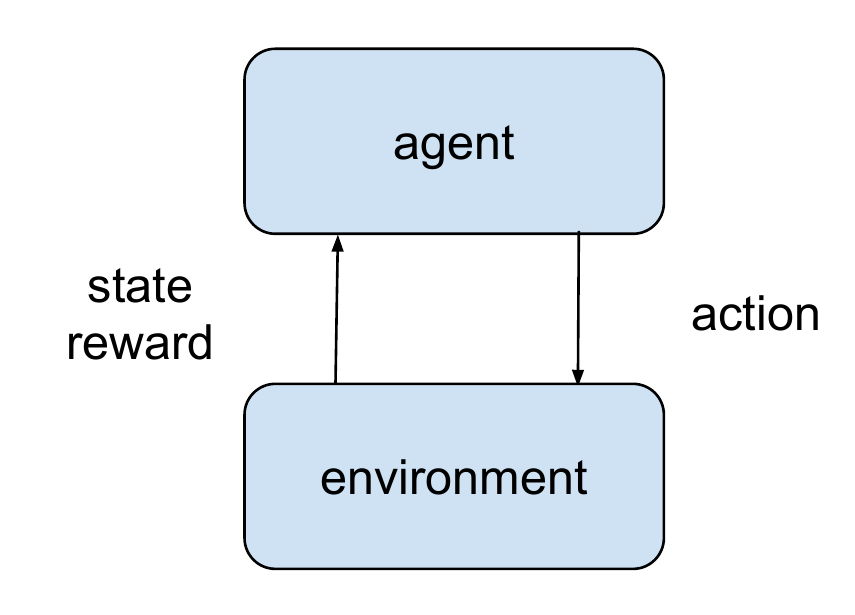}
\caption{Agent-environment interaction.}
\label{agent}
\end{figure}

An RL agent interacts with the environment over time, and learns an optimal policy, by trial and error, for sequential decision-making problems, in a wide range of areas in natural sciences, social sciences, engineering, and art. Figure~\ref{agent} illustrates the agent-environment interaction.

At each time step, an agent receives a state and selects an action, following a policy, which is the agent's behaviour, i.e., a mapping from a state to actions. 
The agent receives a scalar reward and transitions to the next state according to the environment dynamics. 
In an episodic environment, this process continues until the agent reaches a terminal state and then it restarts.
Otherwise, the environment is continuing without a terminal state.
There is a discount factor to measure the influence of future award.
The model refers to the transition probability and the reward function. 
The RL formulation is very general:
state and action spaces can be discrete or continuous;
and an RL problem can be deterministic, stochastic, dynamic, or adversarial as in some games.

A state or action value function measures the goodness of each state or state action pair, respectively.
It is a prediction of the return, or the expected, accumulative, discounted, future reward.
The action value function is usually called the $Q$ function.
An optimal value is the best value achievable by any policy, and the corresponding policy is an optimal policy.
An optimal value function encodes global optimal information, 
i.e., it is not hard to find an optimal policy based on an optimal state value function, 
and it is straightforward to find an optimal policy with an optimal action value function.  
The agent aims to maximize the expectation of a long-term return or to find an optimal policy.

\subsection{More RL Basics}

When the system model is available, we may use dynamic programming methods: 
policy evaluation to calculate value/action value function for a policy, 
and value iteration or policy iteration for finding an optimal policy, 
where policy iteration consists of policy evaluation and policy improvement. 
When there is no model, we resort to RL methods; i.e., RL methods can work without knowing the dynamics of the system.
RL can work with interactions in an online manner.
RL methods also work when the model is available. 
Monte Carlo (MC) methods learn from complete episodes of experience, not assuming knowledge of transition nor reward models, and use sample means for estimation. 
Monte Carlo methods are applicable only to episodic tasks.
%An RL environment can be a multi-armed bandit, an MDP, a partially observable MDP (POMDP), a game, etc.

Temporal difference (TD) learning is central in RL. 
TD learning usually refers to the learning methods for value function evaluation discovered by Richard Sutton in 1988. 
TD learning learns state value function directly from experience, with bootstrapping from its own estimation, in a model-free, online, and fully incremental way.
%SARSA, representing state, action, reward, (next) state, (next) action, is an on-policy control method to find an optimal policy.
 Q-learning is a temporal difference control method, learning an optimal action value function to find an optimal policy.
Q-learning is an off-policy method, learning with experience trajectories from some behaviour policy, but not necessarily from the target policy. 

TD learning and Q-learning are value-based methods.
In contrast, policy-based methods optimize the policy directly, e.g., policy gradient.
Actor-critic algorithms update both the value function and the policy.

In tabular cases, a value function and a policy are stored in  tabular forms. 
Function approximation is a way for generalization when the state and/or action spaces are large or continuous. 
Function approximation aims to generalize from examples of a function to construct an approximate of the entire function.
It is a concept studied in machine learning.
%Function approximation in RL usually treats each backup as a training example, and encounters new issues like nonstationarity, bootstrapping, and delayed targets. 
Linear function approximation is a popular choice, partially due to its desirable theoretical properties. 
A function is approximated by a linear combination of basis functions, which usually need to be designed manually.
The coefficients, or weights, in the linear combination, need to be found by learning algorithms.

We may also have non-linear function approximation, in particular, with DNNs.
We obtain deep RL methods when we integrate deep learning with RL, to represent the state or observation and/or actions, to approximate value function, policy, and model (state transition function and reward function), etc. 
Here, the weights in DNNs need to be found.
Deep RL is popular and has achieved stunning achievements recently, although it dates back a long time ago.
There are several popular deep RL algorithms, like deep Q-network (DQN), 
Asynchronous Advantage Actor-Critic (A3C), 
Deep Deterministic Policy Gradient (DDPG),
Trust Region Policy Optimization (TRPO), 
Proximal Policy Optimization (PPO),
and soft actor-critic.

A fundamental dilemma in RL is exploration vs. exploitation.
An RL agent needs to trade off between exploiting the current best policy and exploring uncertain policies.
The current best policy may be a sub-optimal policy.
A simple exploration approach is $\epsilon$-greedy, 
in which an agent selects a greedy action with probability $1-\epsilon$, and a random action otherwise.
Upper Confidence Bound (UCB) is another exploration method, considering both the action value and its estimation variance.
UCB plays an important role in AlphaGo.

\subsection{RL Parlance}

We explain some terms in RL parlance collectively here to make it convenient for readers to check them. This section is somewhat technical.

The prediction problem, or policy evaluation, is to compute the state or action value function for a policy. The control problem is to find the optimal policy. Planning constructs a value function or a policy with a model.

We have a behaviour policy to generate samples, and we want to evaluate a target policy.
On-policy methods evaluate the behaviour policy with samples from the same policy, e.g., TD learning fits the value function to the current policy, i.e., TD learning evaluates the policy based on samples from the same policy. 
In off-policy methods, an agent learns a value function/policy, maybe following an unrelated behaviour policy. 
For instance, Q-learning attempts to find action values for the optimal policy directly, not necessarily fitting to the policy generating the data, i.e., the policy Q-learning obtains is usually different from the policy that generates the samples. 
The notion of on-policy and off-policy can be understood as same-policy and different-policy, respectively.

The exploration vs. exploitation dilemma is about the agent needing to exploit the currently best action to maximize rewards greedily, yet it has to explore the environment to find better actions, when the policy is not optimal yet, or the system is non-stationary.

In model-free methods, the agent learns with trail-and-error from experience directly; the model, usually the state transition, is not known. RL methods that use models are model-based methods; the model may be given, e.g., in the game of computer Go, or learned from experience.

In an online mode, algorithms are trained on data acquired in sequence. 
In an offline mode, or a batch mode, algorithms are trained on a collection of data.

With bootstrapping, an estimate of state or action value is updated from subsequent estimates.

\subsection{A Shortest Path Example}

Consider the shortest path problem as an example. 
The single-source shortest path problem is to find the shortest path between a pair of nodes to minimize the distance of the path, or the sum of weights of edges in the path. 
%The single-source shortest path problem is to find the shortest path between a pair of nodes to minimize the sum of weights of edges in the path. 
We can formulate the shortest path problem as an RL problem.
The state is the current node. 
At each node, following the link to each neighbour is an action. 
The transition model indicates that the agent goes to a neighbour after choosing a link to follow. 
The reward is then the negative of link cost. 
The discount factor can be 1, so that we do not differentiate immediate reward and future reward. 
It is fine since it is an episodic problem.
The goal is to find a path to maximize the negative of the total cost, i.e., to minimize the total distance. 
An optimal policy is to choose the best neighbour to traverse to achieve the shortest path; 
and, for each state/node, an optimal value is the negative of the shortest distance from that node to the destination. 

\begin{figure}
\centering
\includegraphics[width=0.75\linewidth]{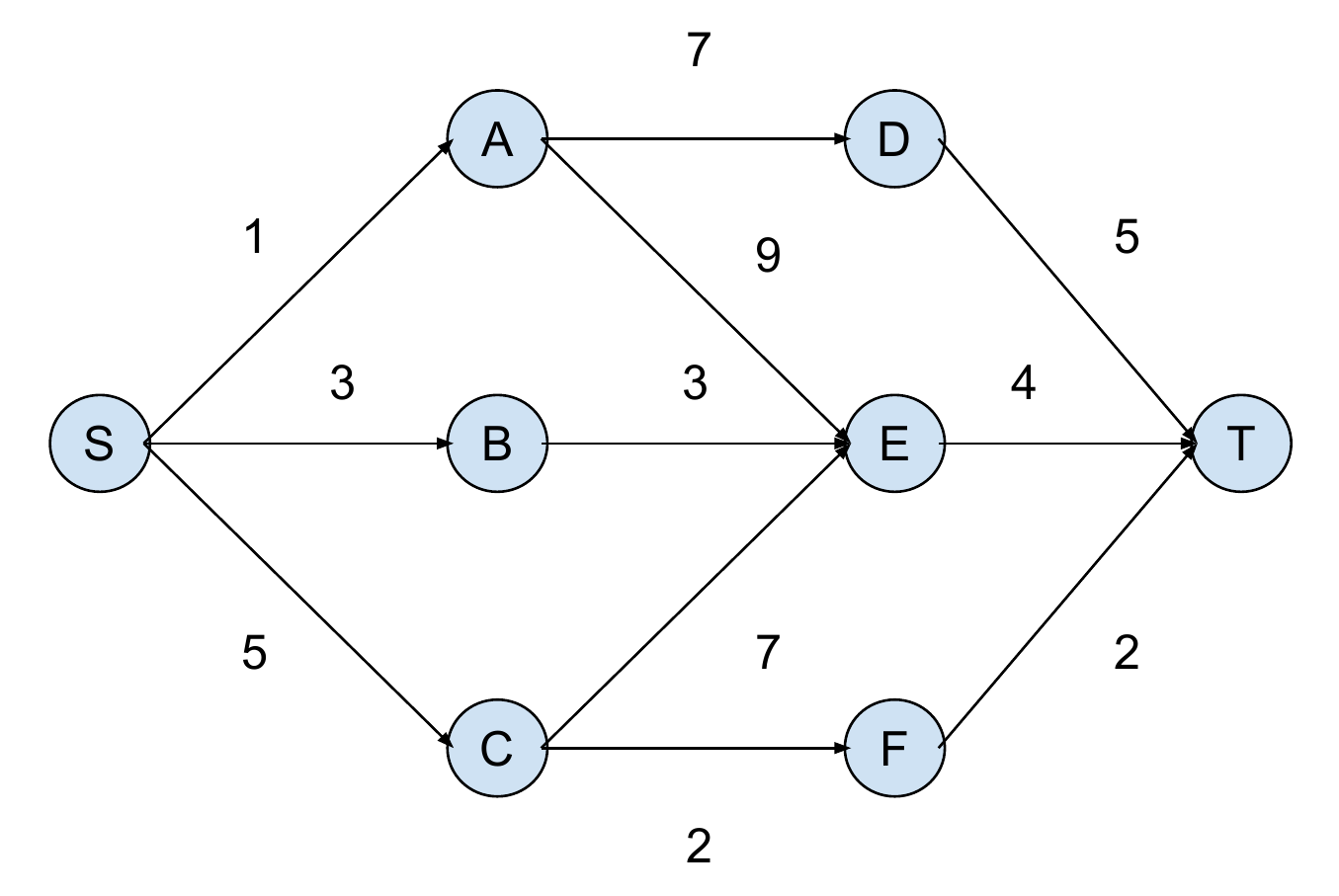}
\caption{An shortest path example.}
\label{shortestpath}
\end{figure}

An example is shown in Figure~\ref{shortestpath}, 
with the graph nodes, (directed) links, and costs on links.
We want to find the shortest path from node $S$ to node $T$.
We can see that if we choose the nearest neighbour of node $S$, i.e., node $A$, as the next node to traverse, then we cannot find the shortest path, i.e., $S \rightarrow C \rightarrow F \rightarrow T$.
This shows that a short-sighted choice, e.g., choosing node $A$ to traverse from node $S$, may lead to a sub-optimal solution.
RL methods, like TD-learning and Q-learning, can find the optimal solution by considering long-term rewards.

Some readers may ask why not use Dijkstra's algorithm.
Dijkstra's algorithm is an efficient algorithm, with the global information of the graph, including nodes, edges, and weights. 
RL can work without such global graph information by wandering in the graph according to some policy, i.e., in a model-free approach. 
With such graph information, Dijkstra's algorithm is efficient for the particular shortest path problem.
However, RL is more general than Dijkstra's algorithm, applicable to many problems. 

\subsection{RL Issues}

Although RL has made significant achievements, there are still many issues. 
RL with function approximation, in particular with DNNs, encounters the deadly triad, i.e., instability and/or divergence caused by the integration of off-policy, function approximation, and bootstrapping.
Sample efficiency, sparse reward, credit assignment, exploration vs. exploitation, and representation are common issues.
Reproducibility is an issue for deep RL, i.e., experimental results are influenced by hyperparameters, including network architecture and reward scale, random seeds and trials, environments, and codebases.
Reward specification may cause problems, and a reward function may not represent the intention of the designer.
An old example is that King Midas wished everything he touched turned into gold. 
Unfortunately, this is not his intention: he did not mean to turn food, family members, etc. into gold.
RL also shares issues with machine learning like time/space efficiency, accuracy, interpretability, safety, scalability, robustness, simplicity, etc.
Take it positively, there are efforts to address all these issues.
We will discuss more in later chapters.

Bearing in mind that there are many issues, RL is an effective technique for many applications.
As discussed by Dimitri Bertsekas, a prominent researcher working on RL, 
one one hand, there are no methods that are guaranteed to work for all or even most problems; 
on the other hand, 
there are enough methods to try with a reasonable chance of success for most types of optimization problems: deterministic, stochastic, or dynamic ones, discrete or continuous ones, games, etc.
He is cautiously positive about RL applications: \enquote{We can begin to address practical problems of unimaginable difficulty!} and \enquote{There is an exciting journey ahead!}

\subsection{Reinforcement Learning for Real Life}

We organized the ICML 2019 Workshop on Reinforcement Learning for Real Life  to bring together researchers and practitioners from industry and academia interested in addressing practical and/or theoretical issues in applying RL to real life scenarios.

We had three superb invited talks:
\begin{itemize}
\item AlphaStar: Mastering the Game of StarCraft II by David Silver
\item How Do We Make Real World Reinforcement Learning Revolution? by John Langford
\item Reinforcement Learning in Recommender Systems: Some Challenges by Craig Boutilier
\end{itemize}

We had a great lineup of panelists:
Craig Boutilier (Google Research),
Emma Brunskill (Stanford),
Chelsea Finn (Google Brain, Stanford, UC Berkeley),
Mohammad Ghavamzadeh (Facebook AI),
John Langford (Microsoft Research),
David Silver (Deepmind), and
Peter Stone (UT Austin, Cogitai).
And important questions were discussed, e.g., 
What application area(s) are most promising for RL?
What are general principles for applying RL to real life applications?

We had around 60 posters/papers. 
We selected four contributed papers as best papers:
\begin{itemize}
\item Chow et al. about RL safety in continuous action problems; %\cite{Chow2019RL4RealLife}
\item Dulac-Arnold et al. about nine challenges of real-life RL; % \cite{Dulac-Arnold2019RL4RealLife}
\item Gauci et al. about Horizon, Facebook's open source applied RL platform; % \cite{Gauci2019RL4RealLife}
\item Mao et al. about Park, an open platform for learning augmented computer systems. %\cite{Mao2019RL4RealLife}
\end{itemize}

Check the workshop website for more details, like videos for invited talks, most papers and some posters, at \url{https://sites.google.com/view/RL4RealLife}.

\subsection{How to Use RL?}

To apply RL to real life applications, first we need to have an RL problem formulation, 
so that we have proper definitions of the environment, the agent, states, actions, and rewards. 
We may have the transition model sometimes.
We may check if supervised learning or contextual bandits are more suitable for the problem;
in that case, RL is not the right formulation.

Sufficient data are essential for the success of current RL techniques. 
Such big data may come from a perfect model, a high-fidelity simulator, or experience.

%RL algorithms are usually data-hungry.
A model or a good simulator can help generate sufficient data for training. 
For some problems, e.g., healthcare, education, and autonomous driving, it may be difficult, infeasible, or unethical to collect data for all scenarios.
In this case, we may have to rely on off-policy learning techniques to learn a target policy from data generated from a different behaviour policy.
There are recent efforts for transferring learning from simulation to realistic scenarios, in particular, in robotics.
Some problems may  require a huge amount of computation.
For example, the success of AlphaGo hinges on two important factors among others: a perfect model (based on game rules) and the Google-scale computation to generate sufficient data from the model for training.
%On the other hand, some problems are not computationally intensive.

With data, we can start feature engineering, which may require significant manual effort and domain knowledge.
In the en-to-end paradigm, minimal or no feature engineering may be required;
however, in practical problems, feature engineering would probably be inevitable and play a critical role in achieving decent performance.

We next consider how to represent the problem, e.g., whether and what kind of neural network to represent the value function and/or the policy. We may also consider linear function approximation. For problems of medium size or smaller, we may even use tabular representation. 

With data, features, and representation, we choose which RL algorithm(s) to find the optimal value/policy.
There are many RL algorithms to choose from, online or offline, on- or off-policy, model-free or model-based, etc.
We usually try several algorithms and choose the best one(s), on a case by case basis.

We then conduct experiments to tune hyperparameters and to examine performance metrics.
It is desirable to compare RL with the state of the art, which may be reinforcement learning, supervised learning, contextual bandit, or some traditional algorithms.
We may iterate the above steps to make improvements.

Once we have decent performance, we will deploy the trained RL model in the real system.
We monitor the performance, and may refine the model periodically. 
We may iterate the above steps to improve the system. 

\begin{figure}
\centering
\includegraphics[width=0.35\linewidth]{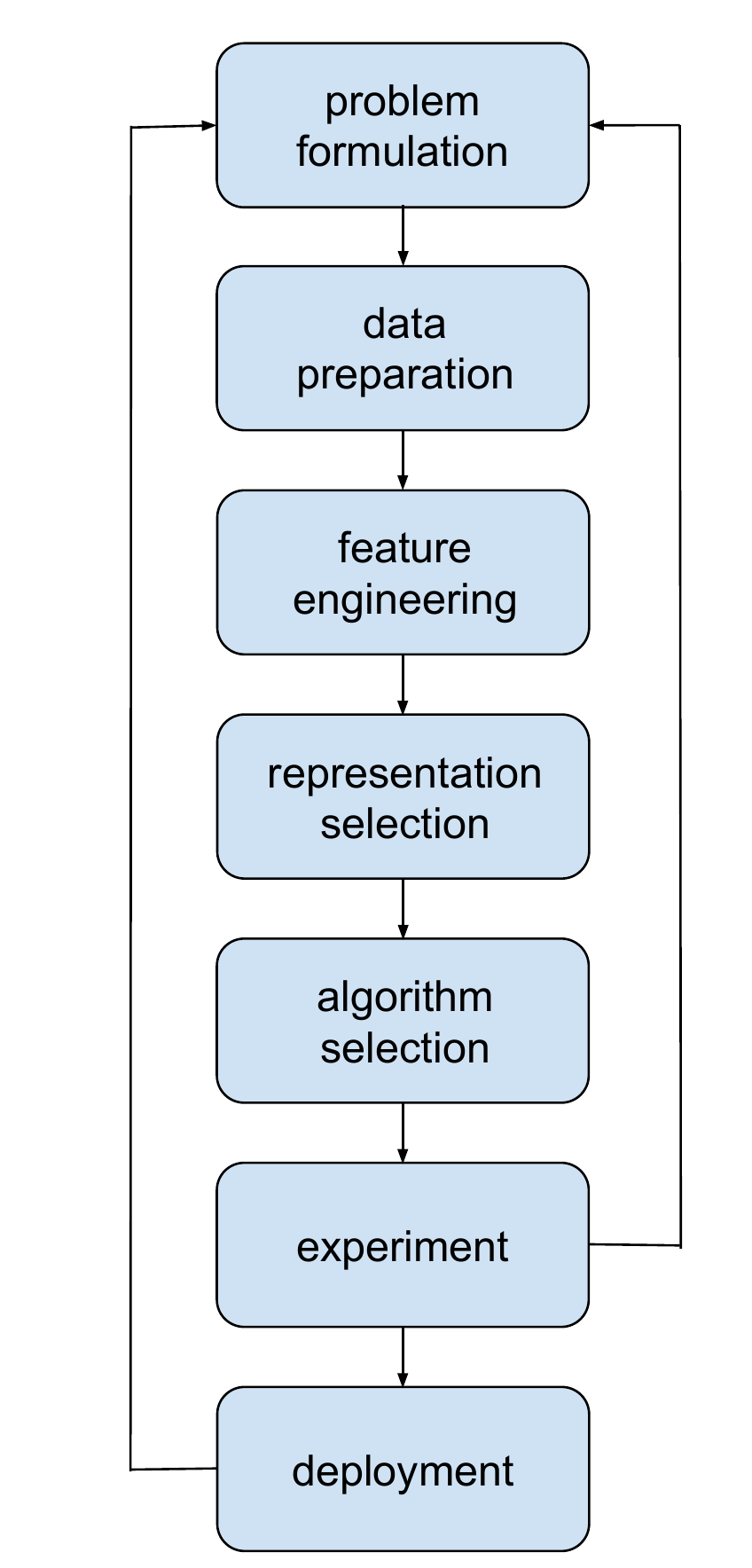}
\caption{The procedure of how to use RL.}
\label{procedure}
\end{figure}

We illustrate how to use RL in Figure~\ref{procedure}, and summarize it concisely as below.
\begin{itemize}
\item Step 1: Formulate the RL problem. Define environment, agent, states, actions, and rewards. 
\item Step 2: Prepare data, from interactions with the environment, and/or a model/simulator.
\item Step 3: Feature engineering, probably manually with domain knowledge. 
\item Step 4: Choose the representation, deep learning or not, non-linear or linear. 
\item Step 5: Choose RL algorithms, online/offline, on-/off-policy, model-free/-based, etc.
\item Step 6: Conduct experiments and may iterate Step 1 - 5 for refinements.
\item Step 7: Deployment. We will monitor and refine the deployed system online or offline. We may iterate Step 1 - 6 for refinement.
\end{itemize}

\subsection{Study Material}

Among RL study material, David Silver's course is classic, OpenAI Spinning Up Deep RL is concise, DeepMind \& UCL's course is new, UC Berkeley Deep RL course is advanced, and Sutton \& Barto's RL book is a must read.

\begin{itemize}
\item David Silver's RL course at UCL
\item[] \url{http://www0.cs.ucl.ac.uk/staff/D.Silver/web/Teaching.html}

\item OpenAI Spinning Up Deep RL
\item[] \url{https://blog.openai.com/spinning-up-in-deep-rl/}

\item DeepMind \& UCL Advanced Deep Learning and Reinforcement Learning Course,
\item[] \url{https://bit.ly/2KwA5gi}
%\item[] \url{https://www.youtube.com/playlist?list=PLqYmG7hTraZDNJre23vqCGIVpfZ_K2RZs}

\item UC Berkeley Deep RL course
\item[] \url{http://rail.eecs.berkeley.edu/deeprlcourse/}

\item Sutton \& Barto's RL book
\item[] \url{http://www.incompleteideas.net/book/the-book-2nd.html}
\end{itemize}

It is essential to understand concepts in deep learning and machine learning. Here are several excellent overview papers.
\begin{itemize}
\item LeCun, Bengio and Hinton, Deep Learning, Nature, May 2015
\item Jordan and Mitchell, Machine learning: Trends, perspectives, and prospects, Science, July 2015
\item Michael Littman, Reinforcement learning improves behaviour from evaluative feedback, Nature, May 2015
\end{itemize}

%https://lilianweng.github.io/lil-log/2018/02/19/a-long-peek-into-reinforcement-learning.html

It is essential to test the understanding of concepts with coding.
OpenAI gym is popular, \url{https://gym.openai.com}.
This Github open source has implementations of examples in Sutton \& Barto's RL book, and many deep RL algorithms,
\url{https://github.com/ShangtongZhang/reinforcement-learning-an-introduction}.

This blog is a big collection of 
resources for deep RL,
\url{https://medium.com/@yuxili/resources-for-deep-reinforcement-learning-a5fdf2dc730f}.

Here are some blogs,
\url{https://medium.com/@yuxili}

\subsection{The Time for Reinforcement Learning is Coming}

RL is a general learning, predicting, and decision making paradigm. 
RL is probably helpful for an application, if it can be regarded as or transformed into a sequential decision-making problem, and states, actions, maybe rewards, can be constructed. 
There is a chance for RL to help automate and optimize a manually designed strategy in a task.

RL is sequential, far-sighted, and considers long-term accumulative rewards, whereas, 
supervised learning is usually one-shot, myopic, and considers instant rewards.
Such long-term consideration is critical for achieving optimal strategies for many problems.
In the shortest path problem, for instance, if we consider only the nearest neighbour at a node, we may not find the shortest path.

David Silver, the major researcher behind AlphaGo, gave a hypothesis: AI = RL + DL. 
We see some supporting evidence.
Russell and Norvig's AI textbook states that 
\enquote{reinforcement learning might be considered to encompass all of AI: an agent is placed in an environment and must learn to behave successfully therein} and 
\enquote{reinforcement learning can be viewed as a microcosm for the entire AI problem}. 
It is also shown that tasks with computable descriptions in computer science can be formulated as RL problems. 

\begin{figure}
\centering
\includegraphics[width=0.8\linewidth]{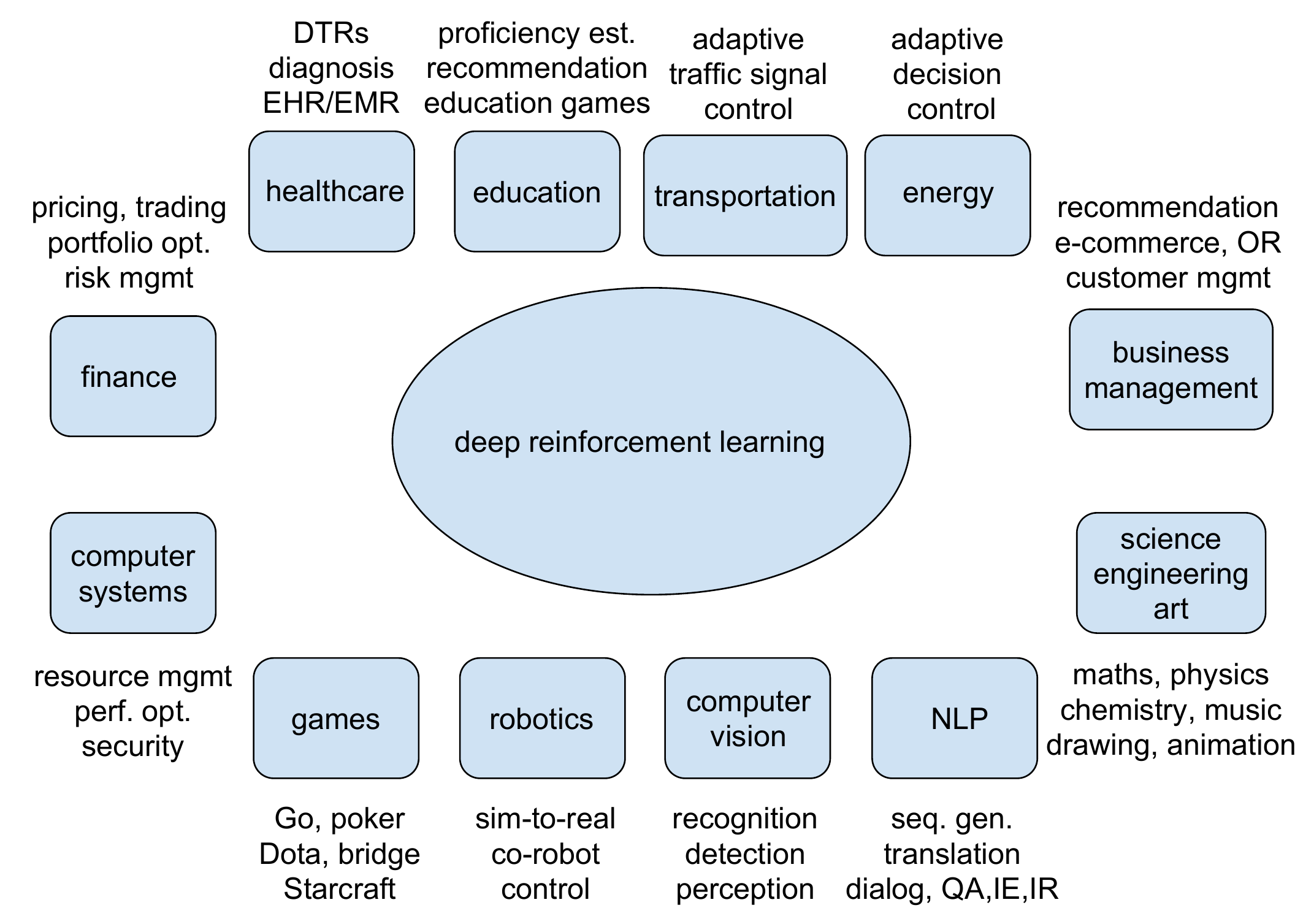}
\caption{RL applications.}
\label{app}
\end{figure}

DL and RL were selected as one of the MIT Technology Review 10 Breakthrough Technologies in 2013 and 2017, respectively. 
DL has been enjoying wide applications.
RL will play more and more important roles in real life applications.
Figure~\ref{app} shows potential applications of RL.
We have witnessed applications of RL in some areas like recommender systems, ads, and probably in finance.
It is probably the right time to nurture, educate, and lead the market for RL. 
We will see both DL and RL prospering in the coming years and beyond.

\subsection{Book Outline}

In the rest of the book, we will discuss several application areas of RL: 
\begin{itemize}
\item recommender systems in Chapter 2, 
  discussing Decision Service, 
  Horizon: an open source applied RL platform,
  news recommendation,
  and multiple items recommendation;
\item computer systems in Chapter 3,  
  discussing neural architecture search,
  device placement,
  data augmentation,
  cluster scheduling,
  an open platform for computer systems
  and NP-hard problems;
\item energy in Chapter 4, discussing smart grid and data center cooling; 
\item finance in Chapter 5, discussing option pricing and order book execution;
\item healthcare in Chapter 6, discussing dynamic treatment strategies and medical image report generation;
\item robotics in Chapter 7, discussing dexterous robot and legged robot;  
\item transportation in Chapter 8, discussing ride-sharing order dispatching.  
\end{itemize}
%We close the book with discussions in Chapter 10.

We do not list games above, a popular testbed/application for RL algorithms.
We discuss computer vision and natural language processing (NLP) when necessary, 
which are fundamental AI techniques to applications, although they can benefit from RL themselves too.
At the end of each chapter, we present an annotated bibliography to discuss relevant reference.

We hope to write a book for managers, investors, researchers, developers, students, and general public interested in AI, and in particular, in RL.
We do not assume the readers to have background in RL, 
although decent RL background is helpful for understanding RL applications.
We hope to convey that RL is very promising and has generated or will soon generate revenues in many applications.

\subsection{Annotated Bibliography}

\citet{Sutton2018} is an intuitive, definite textbook for RL.
\citet{Bertsekas96} discuss neuro-dynamic programming, a theoretical treatment of RL.
\citet{Szepesvari2010} discuss RL algorithms.
\citet{Bertsekas2019} introduce RL and optimal control.
\citet{Powell11} discusses approximate dynamic programming, with applications in operations research.
RL, approximate dynamic programming, and optimal control overlap with each other significantly.
\citet{Goodfellow2016} is an introduction to deep learning.
\citet{Russell2009} is an introduction to AI.

\citet{Mnih-DQN-2015} introduce Deep Q-Network (DQN).
\citet{Silver-AlphaGo-2016} introduce AlphaGo. 
\citet{Silver-AlphaGo-2017} introduce AlphaGo Zero, mastering the game of Go without human knowledge.
\citet{Silver-AlphaZero-2018} introduce AlphaZero, extending AlphaGo Zero to more games like chess and shogi (Japanese chess).
\citet{Moravcik2017} introduce DeepStack and \citet{Brown2017Science} introduce Libratus, for no-limit, heads-up Texas Hold'em Poker.
 
AlphaStar defeats top human players at StarCraft II.\footnote{\url{https://deepmind.com/blog/alphastar-mastering-real-time-strategy-game-starcraft-ii/}}
\cite{Jaderberg2018Quake} present human-level performance at Catch The Flag.
OpenAI Five defeated good human players at Dota 2.\footnote{\url{https://blog.openai.com/openai-five/}}
Such achievements in multi-player games show the progress in mastering tactical and strategical team plays. 
OpenAI trained Dactyl for a human-like robot hand to dextrously manipulate physical objects.\footnote{\url{https://blog.openai.com/learning-dexterity/}}
\cite{Hwangbo2019} present  agile and dynamic motor skills for legged robots.
\cite{Peng2018} present DeepMimic for simulated humanoid to perform highly dynamic and acrobatic skills. 
%Learning to dress achieves dressing tasks with a cloth simulation model. 
\citet{Lazic2018NIPS} study data center cooling with model-predictive control (MPC).
\citet{Segler2018} apply RL to chemical retrosynthesis.
\citet{Popova2018} apply deep RL to computational de novo drug design.

DQN integrates Q-learning with DNNs, and utilizes the experience replay and a target network to stabilize the learning.
In experience replay, experience or observation sequences, i.e., sequences of state, action, reward, and next state, are stored in the replay buffer, and sampled randomly for training.
A target network keeps its separate network parameters, which are for the training, and updates them only periodically, rather than for every training iteration.
\cite{Mnih-A3C-2016} present Asynchronous Advantage Actor-Critic (A3C), in which parallel actors employ different exploration policies to stabilize training, and the experience replay is not utilized.
Deterministic policy gradient can help estimate policy gradients more efficiently.
\cite{Silver-DPG-2014} present Deterministic Policy Gradient (DPG) and \citet{Lillicrap2016}  extend it to Deep DPG (DDPG).
Trust region methods are an approach to stabilize policy optimization by constraining gradient updates.
\cite{Schulman2015} present Trust Region Policy Optimization (TRPO) and 
\cite{Schulman2017PPO} present Proximal Policy Optimization (PPO).
\cite{Haarnoja2018} present soft actor-critic.

%\cite{LiuYao2018NIPS}

%\cite{Jiang2016Doubly}

It is worthwhile to monitor researchers like  Pieter Abbeel, Dimitri Bertsekas, Emma Brunskill, Chelsea Finn, Leslie Kaelbling, Lihong Li, Michael Littman,  Joelle Pineau, Doina Precup, Juergen Schmidhuber, David Silver, Satinder Singh, Peter Stone,  Rich Sutton, Csaba Szepesv{\'a}ri,  and institutes like  CMU, Deepmind,  Facebook,  Google, Microsoft, MIT, OpenAI, Stanford, University of Alberta, UC Berkeley, among many others.

\cite{Amershi2019ICSE} discuss software engineering for machine learning, which would also be helpful for reinforcement learning.
The authors illustrate nine stages of the machine learning workflow, namely, model requirements, data collection, data cleaning, data labeling, feature engineering, model training, model evaluation, model deployment, and model monitoring.
There are many feedback loops in the workflow, e.g., between model training and feature engineering,
and model evaluation and model monitoring may loop back to any previous stages.
The authors also identify three major differences in software engineering for AI from that for previous software application domains:
1) it is more complex and more difficult to discover, manage, and version data;
2) different skill set for model customization and model reuse; and
3) AI components are less modular and more entangled in complex ways.

\cite{Li2017DeepRL} discusses deep RL in an overview style,
drawing a big picture of deep RL, filled with details, focusing on contemporary work, and in historical contexts.
It discusses 6 core elements: value function, policy, reward, model, exploration vs. exploitation, and representation;
6 important mechanisms: attention and memory, unsupervised learning, hierarchical RL, multi-agent RL, relational RL, and learning to learn; 
and 12 applications: games, robotics, natural language processing (NLP), computer vision, finance, business management, healthcare, education, energy, transportation, computer systems, and, science, engineering, and art.

\clearpage

\section{Recommender Systems}
\label{recommender}

A recommender system can suggest products, services, information, etc. to users according to their preferences. It helps alleviate the issues with overwhelming information these days by personalized recommendations, e.g., for news, movies, music, restaurants, etc. 

In user-centric recommender systems, 
it is essential to understand and meet genuine user needs and preferences, 
using natural, unobtrusive, transparent interaction with the users.   
Recommender systems need to
estimate, elicit, react to, and influence user latent state, e.g., satisfaction, transient vs. persistent preferences, needs, interests, patterns of activity etc.
by natural interaction with users,
and acting in user's best interests, i.e, user preferences, behavioural processes governing user decision making and reasoning, etc.
RL will play a central role in all of these.

User-facing RL encounters some challenges,
1) large scale with massive number of users and actions, multi-user (MDPs), and combinatorial action space, e.g., slates;
2) idiosyncratic nature of actions, like stochastic actions sets, dynamic slate sizes, etc.;
3) user latent state, leading to high degree of unobservability and stochasticity, w.r.t. preferences, activities, personality, user learning, exogenous factors, behavioral effects, etc, which add up to long-horizon, low signal noise ratio (SNR) POMDPs; 
and,
4) eco-system effects, e.g., incentives, ecosystem dynamics, fairness, etc.

%RL has wide applications in business, e.g., recommender systems, customer management, mechanism design, etc. In the following we introduce using RL for recommender systems.

\subsection{Decision Service}

Contextual bandits are effective to select actions based on contextual information, 
e.g., in news article recommendation, 
selecting articles for users based on contextual information of users and articles, 
like historical activities of users and descriptive information and categories of content.
This framework covers a large number of applications, 
and Table~\ref{DecisionServiceTable} shows some examples.

\begin{table}[h]
\centering
%\resizebox{\textwidth}{!}{%
\begin{tabular}{ |c|c|c|c|}
\hline

& \makecell{news website\\ (News)} & \makecell{tech support assistance\\ (TechSupp)} & \makecell{could controller\\ (Cloud)} \\ \hline
\hline

\makecell{\textbf{decision}\\\textbf{to optimize}} & \makecell{article to display\\ on top} & \makecell{response\\ to query} & \makecell{wait time\\ before reboot\\ unresponsive machine} \\ \hline
\textbf{context} & \makecell{location, \\browsing history, ...} & \makecell{previous \\dialog elements} & \makecell{machine hardware/OS,\\ failure history, ...} \\ \hline
\makecell{\textbf{feasible}\\\textbf{actions}} & \makecell{available \\news articles} & \makecell{pointers to \\potential solutions} & \makecell{minutes\\ in \{1, 2, ..., 9\}} \\ \hline
\textbf{reward} & \makecell{click/no-click} & \makecell{(negative) human\\ intervention request} & \makecell{(negative)\\ total downtime}  \\ \hline
\end{tabular}
%}
\caption{Example applications of the Decision Service. (reproduced from \cite{Agarwal2016})}
\label{DecisionServiceTable}
\end{table}

However, such setting encounters two challenges.
1) Partial feedback. The reward is for the selected action only, but not for unexplored actions.
2) Delayed rewards. Reward information may arrive considerably later after taking an action.
As a result, an implementation usually faces the following failures:
1) partial feedback and bias;
2) incorrect data collection; 
3) changes in environments; and
4) weak monitoring and debugging.

A/B testing is an approach for exploration to address the issue of partial feedback.
It tests two policies by running them on random user traffic proportionally.
Thus the data requirements scale linearly with the number of policies.
Contextual bandits allows to test and optimize exponential number of policies with the same amount of data.
And it does not require policies to be deployed to test, to save business and engineering effort.
Such capacity is referred to as multiworld testing (MWT).
Supervised learning does not support exploration, so it does not tackle contextual settings or the issues of partial feedback and bias.

The Decision Service defines four system abstractions, namely, 
explore to collect data, log data correctly, learn a good model, and deploy the model in the application,
to complete the loop for effective contextual learning,
by providing the capacity of  MWT with techniques of contextual bandits and policy evaluation.

\begin{figure}[h]
\centering
\includegraphics[width=0.8\linewidth]{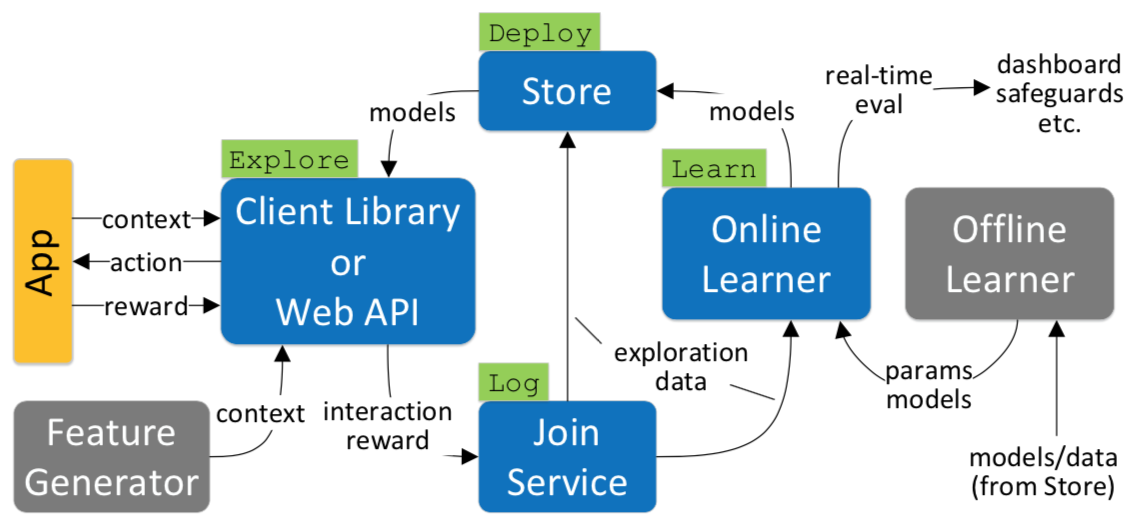}
\caption{DecisionService Architecture (from~\cite{Agarwal2016})}
\label{DecisionService}
\end{figure}

Figure~\ref{DecisionService} shows the Decision Service architecture.
The Client Library implements various exploration policies to implement the Explore abstraction.
It interfaces with the App, takes context feature and event key as input and outputs an action.
It sends data to the Log component, which generates data correctly by the Join Service to log data at the point of decision.
Log sends the exploration data to the Online Learner and the Store.
The Explore and Log components help address failure 1, partial feedback and bias and failure 2, incorrect data collection.
The Online Learner provides online learning from contextual bandit exploration data to implements the Learn abstraction.
It continuously incorporates data and checkpoints models at a configurable rate to the Deploy component.
This helps address failures 3, changes in environments and failure 4, weak monitoring and debugging.
The Store provides model and data storage to implement the Deploy abstraction.
The Offline Learner also uses the data for offline experimentation,
to tune hyperparameters, to evaluate other learning algorithms or policy classes, to change the reward metric, etc.,
in a counterfactually correct way.
The Feature Generator generates features automatically for certain types of content.

The design also helps make the system responsive, reproducible, scalable, fault tolerant, and flexible for deployment.
The Decision Service has several deployment options:
self-hosted in a user's Azure account,
hosted in Microsoft's Azure  account to share resources with other users,
and local model on a user's machine.

We list several applications of Decision Service in the following.
MSN is basically the News problem in Table~\ref{DecisionServiceTable},
where MSN's front-end servers decide how to order articles on the homepage according to a user's requests.
Experiments show $>25\%$ click through rate (CTR) improvements.
Complex recommends videos to come with news articles and recommends top news articles, and achieves a lift of $>30\%$.
TrackRevenue maximizes revenue with recommendation of landing pages for in-app advertisement, and achieves 18\% revenue lift.
Toronto provides answers to technical support questions to reduce the load.
Azure Compute optimizes wait time before rebooting unresponsive virtual machines in a cloud service, and estimates a 19\% reduction of waste time.

\subsection{Horizon: An Open Source Applied RL Platform}

Horizon is Facebook's open source applied RL platform.
It has the following features.

1) Data preprocessing. 
A Spark pipeline converts logged data into the format suitable for training various deep RL models.
In production systems, data are logged as they come, and an offline logic joins the data into the format for RL training.

2) Feature normalization. 
Horizon analyzes the training data and optimizes the transformation function and parameters to normalize features
to learn faster and better.

3) Data understanding tools.
Horizon uses a data-driven, model-based approach to check if several problem properties conform to the MDP formulation,
e.g., if state features are not important to predict next states, then the problem is not sequential,
and, if state features do not predict rewards, then the problem may be formulated as multi-arm bandits.
The model can be used to compute feature importance and select important ones for training.
This provides explainability to RL practitioners.
 
4) Model implementation.
Horizon contains variants of DQN for problems with discrete action spaces, 
parametric action DQN to handle large action spaces,
and DDPG and soft actor-critic for problems with continuous action spaces.

5) Multi-node and multi-GPU training.
Horizon supports training with a CPU, a GPU, multiple GPUs, and multiple nodes.

6) Model understanding and evaluation.
Horizon provides training metrics of temporal difference loss (TD loss) and Monte Carlo loss (MC loss) 
to show the stability and convergence of the training process.
TD loss measures the function approximation error.
MC loss measures the difference between the model's Q-value and the logged value.

Counterfactual policy evaluation (CPE) methods can predict the performance of a policy without deploying it.
Otherwise, many A/B tests may be required to find the optimal model and hyperparameters. 

7) Optimized serving.
Horizon exports trained models from PyTorch to Caffe2 via ONNX. 
Data are logged with unique keys, including
possible actions, 
the propensity of choosing an action,
the selected action, and
the received reward.
The reward usually comes later, and join operations are needed to form complete data points.

8) Tested algorithms.
Horizon borrows ideas from best practices in building software systems and tests algorithms with unit tests and integration tests.

\begin{figure}[h]
\centering
\includegraphics[width=0.5\linewidth]{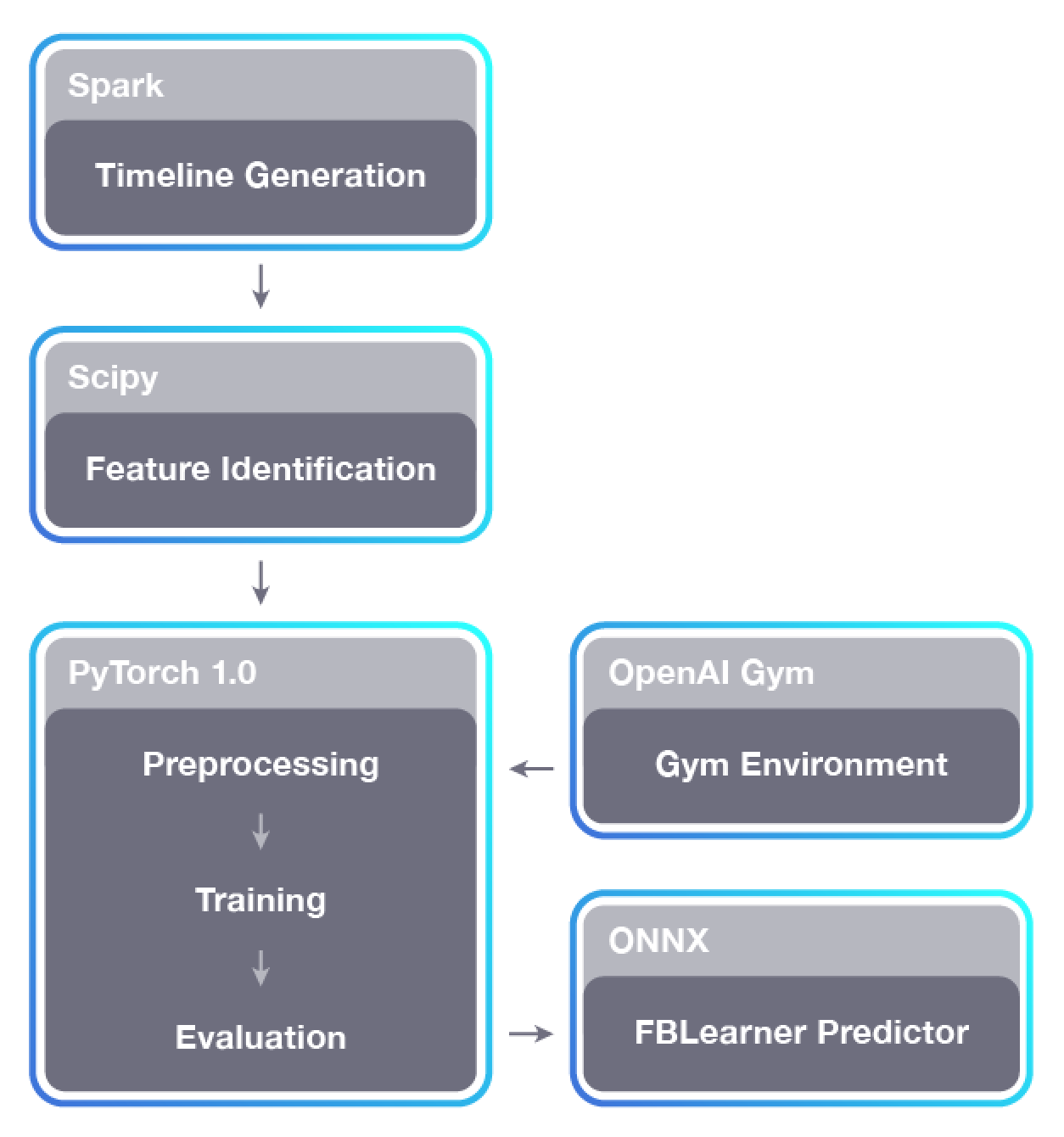}
\caption{Horizon RL pipeline. (from \url{https://code.fb.com/ml-applications/horizon/})}
\label{Horizon}
\end{figure}

The Horizon RL pipeline, as illustrated in Figure~\ref{Horizon}, consists of: 
1) timeline generation, which runs across thousands of CPUs;
2) training, which runs across many GPUs; and
3) serving, which spans thousands of machines.

Horizon has been deployed for notification services for push notifications and page administrator notifications
at Facebook to deliver more relevant notifications, 
and has also been applied to video bitrate adaptation and other applications.

Push notifications are a channel for personalized and time sensitive updates, 
about posts or stories, friends updates, groups, pages, events, etc.
An MDP is formulated as follows.
An action is sending or dropping a notification.
A state consists of features representing the person and the notification candidate.
Rewards reflect interactions and activities on Facebook, and volume of notifications sent.
If the difference between the values of sending and dropping is large, sending the notification has a significant value.
A/B tests show that the tested RL model outperforms the control non-RL model.

\subsection{News Recommendation}

One approach using RL for single item recommender systems follows.
The system may recommend multiple items; however, the recommender systems are not designed for multiple items in mind, e.g., by ordering items according to their Q values.

The model is built for news recommendation; however, the principles applies to general recommender systems.
Several challenges it needs to handle are: 1) difficulties in handling the dynamics in news recommendations, e.g., news becomes obsolete fast, and users change their preferences from time to time; 
2) most recommender systems considering users' click/no click or ratings as feedback, but not including users' return behaviour;
and 3) the diversity of recommended items. 

\begin{figure}[h]
\centering
\includegraphics[width=0.5\linewidth]{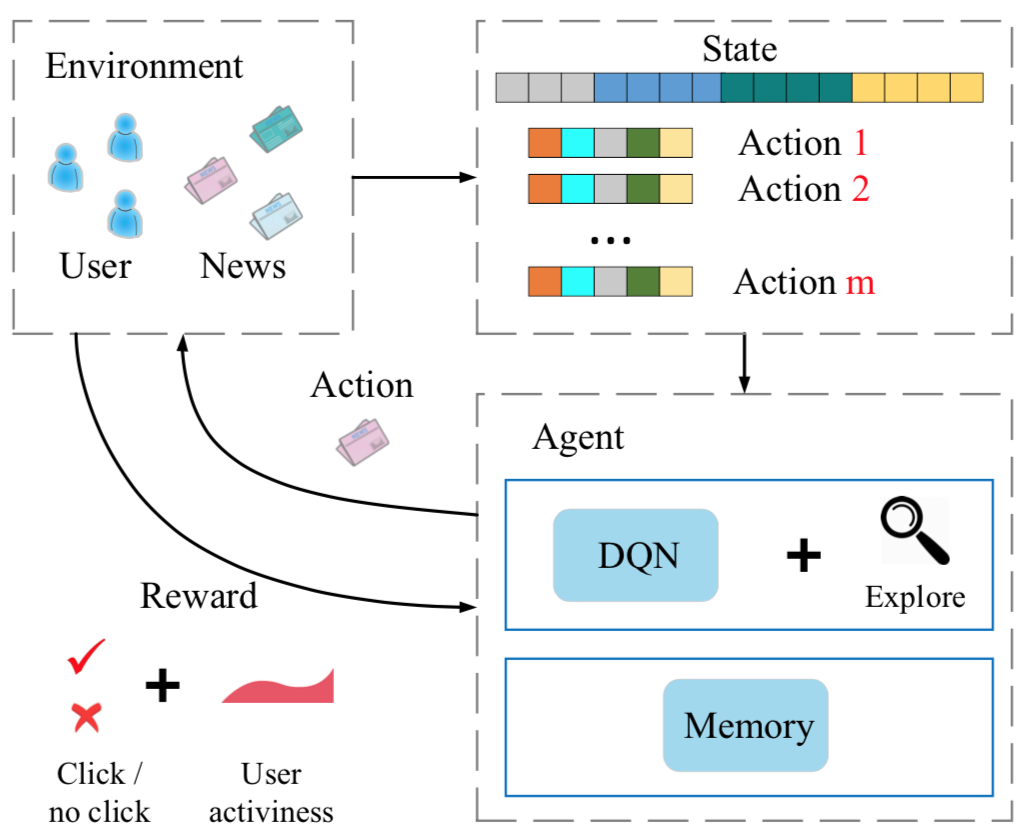}
\caption{Deep RL news recommender system. (from~\cite{Zheng2018WWW})}
\label{DRN1}
\end{figure}

To formulate a recommender system as an RL problem, we need to define the environment, the agent, states, actions, and rewards, as illustrated in Figure~\ref{DRN1}.
Users and news form the environment.
The recommendation algorithm is the agent.

To define states and actions, we first discuss feature construction.
There are four categories of features: news features, user features, user news features, and context features.
News features describe if a piece of news contains certain properties like headline, provider, and topic category over several time granularities respectively.  
The time granularity can be the last 1 hour, 6 hours, 24 hours, 1 week, and 1 year.
User features describe if the user clicks news with certain properties like those in news features over several time granularities respectively.
User news features describe the interaction between users and news, including the frequency of the entity, category, topic category, and provider appearing in users' histories.
Context features describe the context of the occurrence of a news request, including its time, weekday, and freshness of news, i.e., the gap between the publish time and the request time.   
Then, state features are defined with user features and context features, and action features are define with user news features and context features.

Usually, for recommender systems, a reward is defined by the implicit click/no click information or explicit ratings. 
However, some factors, e.g., user activeness, may also play an important role.
User activeness is related to the frequency users checking news and behaviours of users return.
A reward is then defined as a combination of signals for clicks and user activeness.

\begin{figure}[h]
\centering
\includegraphics[width=0.8\linewidth]{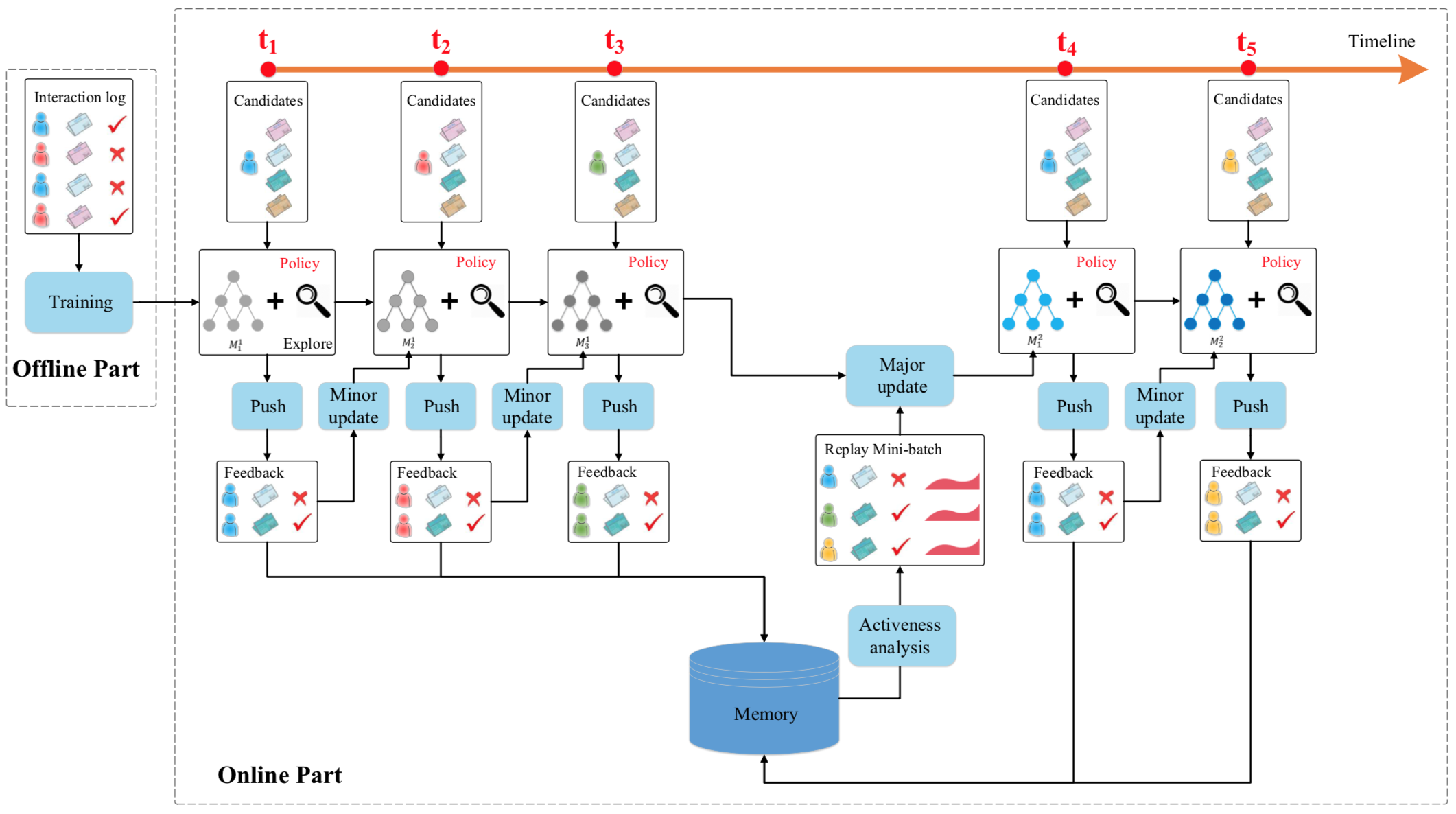}
\caption{Deep RL recommender system model framework. (from~\cite{Zheng2018WWW})}
\label{DRN2}
\end{figure}

Next we discuss the deep RL recommender system model framework, as illustrated in Figure~\ref{DRN2}.

The recommendation model starts with offline handling of the four features, extracting from users and news.
The model is trained offline with user-news click logs using a Deep Q-Network (DQN).
After this, the model enters the online mode.

When the agent receives a news request, it recommends a top-$k$ list of news, based on features of the user and news.
Technically, the agent utilizes the state action value function $Q(s,a)$ to select the most $k$ preferable news for the user.
The user then responds with click or not as a feedback.
The agent may update the decision network $Q$ after each recommendation or after certain time duration, e.g., 1 hour, with many recommendations.
The model repeats the above procedure to refine the decisions.

Recommender systems also face the exploration vs. exploitation dilemma.
The popular methods like $\epsilon$-greedy or upper confidence bound (UCB) make random recommendations,
which may cause undesirable results.
Exploration is also necessary to improve the diversity of the recommended news. 
One method is to explore news candidates close to those the agent would recommend.
 
There are several performance metrics to evaluate a recommender system.
Click Through Rate (CTR) is defined as the number of clicked items divided by the number of total items.
Precision@$k$ is defined as the number of clicks in top-$k$ recommended items divided by $k$.
Normalized Discounted Cumulative Gain (nDCG)  is related to ranks of items in the recommendation list, the length of the list, the ranking function, and the information about click or not. nDCG is basically the sum of discounted  clicks, where discount factors are related to ranks.
One way to define recommendation diversity is the sum of cosine similarity between different recommended item pairs divided by the number of such item pairs. For two items, the more different one is from another, the lower is the cosine similarity in the range of [0,1].

The method compares with several baselines, using logistic regression, factorization machines, wide \& deep neural networks, and linear UCB. The experiments are conducted with offline dataset and with online deployment in a commercial news recommendation application.
The method outperforms the baselines.

\subsection{Multiple Items Recommendation}

% copied from RL4RealLife summary, July 7, 2019; paraphrasing is necessary 

A recommender system needs to recommend multiple items at the same time,
which is also called slate recommendation, e.g., a slate of videos to users visiting YouTube.
In this problem, there may be non-trivial long/short-term trade-offs, and RL is a natural solution.

In the MDP/RL formulation,
states refer to user features, user history, and contextual features;
actions refer to possible slates of recommendations;
reward refers to immediate engagement of a selected recommendation, e.g., watched video;
and the discount factor is chosen to reflect target/expected session length.
The combinatorics of slates pose problems for generalization, exploration and optimization.
An ideal way is to decompose value of slate into value of constituent items.   

We may also encounter the item interaction problem,
so that the presence of some items on the slate impacts user response (hence value) of other items.
As a result, the value of slate depends on user choice model, which requires joint optimization of the slate.

One approach is SlateQ with slate decomposition.
SlateQ decomposes slate Q-values into item Q-values, 
with specific assumptions about user choice from slate,
i.e., user selects one item from slate, and state transitions and reward depend only on selected items.
We will learn conditional-on-click, item-wise Q-values, 
and slate Q-values are derived and fed into item Q-values,
with decomposed TD/SARSA updates and full Q-learning updates.

Slate optimization is tractable for certain user choice models.
The standard formulation is fractional mixed-integer program.
It is a simplified product-line or assortment optimization problem.
It can be relaxed and transformed into linear program to obtain an optimal solution.
We can also apply top-$k$ and greedy methods.

RecSim is a synthetic environment.  
Documents are represented by topic vector and quality score, both of which are random.
Users are represented by interest vector over topics and satisfaction, which are random initially.
User interaction describes that it is more likely for a user to consume an item on the slate if interested.
User state dynamics describes that, 
interests move stochastically in the direction of the consumed item,
satisfaction increases or decreases stochastically with item quality,
and less satisfied scenarios occur with greater odds of terminating session, e.g., due to time budget.
There are tradeoffs between nudging toward higher-quality topics and decreasing immediate engagement.
Experiments are conducted with 20 documents, 10 candidates and 3 items per slate to test full Q-learning. 

\subsection{Annotated Bibliography}

Recommender systems have attracted a lot of attention recently.
Popular approaches include content based methods, collaborative filtering based methods, and hybrid methods.
Recently, deep learning plays an important role.
See \cite{Zhang2017Recommender} for a survey.

The discussion of the Decision Service is based on \cite{Agarwal2016}.
\cite{Li2010} formulate personalized news articles recommendation as a contextual multi-armed bandit problem, 
considering contextual information of users and articles, 
such as historical activities of users and descriptive information and categories of content. 
In addition, \cite{Karampatziakis2019RL4RealLife} discuss lessons from real life RL in a customer support bot.

The discussion about Horizon is based on~\cite{Gauci2019RL4RealLife}
and a blog titled Horizon: The first open source reinforcement learning platform for large-scale products and services at \url{https://code.fb.com/ml-applications/horizon/}.

The discussion about single item recommendation, in particular, news recommendation, is based on \cite{Zheng2018WWW}. 
See \citet{ZhaoXiangyu2018} for a survey on deep RL for search, recommendation, and online advertising.

The discussion of user-facing RL in the beginning and multiple item recommendation is based on the invited talk by Craig Boutilier titled Reinforcement Learning in Recommender Systems: Some Challenges in the ICML 2019 Workshop on Reinforcement Learning for Real Life, at \url{https://sites.google.com/view/RL4RealLife}. 
\cite{Ie2019arXiv} discuss SlateQ. 
\cite{Ie2019RL4RealLife} discuss RecSim.
\cite{Chen2019ICML} and \cite{Shi2019AAAI} also discuss model building.

In addition, 
\cite{Theocharous2015} formulate a personalized ads recommendation systems as an RL problem to maximize life-time value (LTV) with theoretical guarantees; 
\cite{ChenShiYong2018} propose to stabilize online recommendation with stratified random sampling and approximate regretted reward.
\cite{Zhao2018RecSys} study page-wise recommendation;
\cite{Zhao2018KDD} study the effect of not only positive but also negative feedback;
\cite{Hu2018ranking} study ranking in e-commerce search engine.

\clearpage

\section{Computer Systems}
\label{system}

Many problems in computer systems are about sequential decision making,
and manual design is common for strategies involved.
RL is an approach to automate and optimize such strategies.

\subsection{Neural Architecture Search}

Deep learning has significant achievements recently in image processing, speech processing, and natural language processing.
How to design neural network architectures is a critical issue.
Neural architecture search is an approach using RL.

\begin{figure}[h]
\centering
\includegraphics[width=0.7\linewidth]{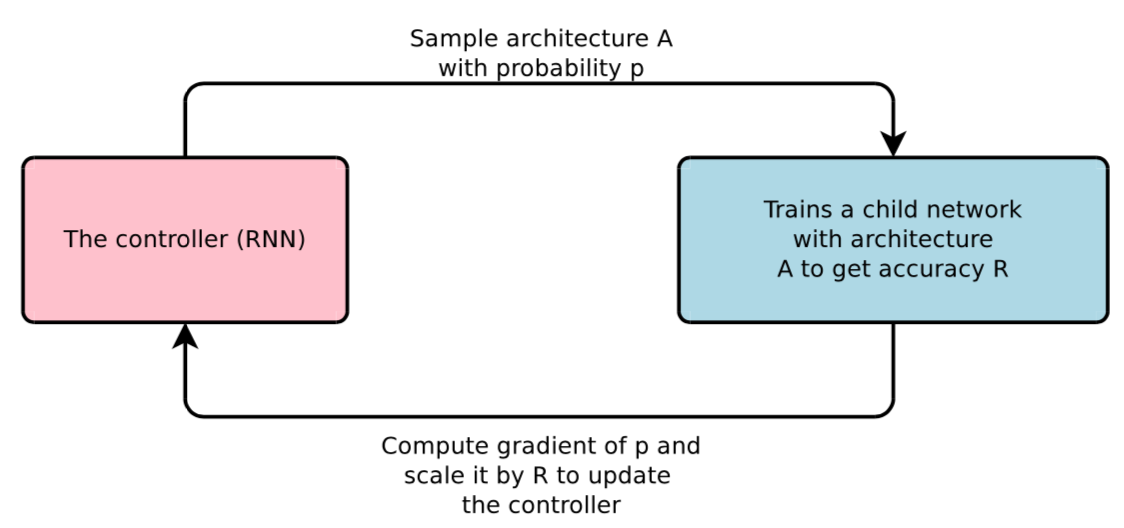}
\caption{An overview of neural architecture search. (from \cite{Zoph2017})}
\label{NAS}
\end{figure}

In neural architecture search, as illustrated in Figure~\ref{NAS},
a recurrent neural network (RNN) is used to generate neural network descriptions and is trained to maximize the expected accuracy on a validation data set with RL. 
A variable-length string represented by the RNN specifies the structure and connectivity of a neural network.
The list of tokens in the string can be regarded as a list of actions to design a neural network architecture.
This can represent different components in neural networks, like  1) for a convolutional layer, including hyper-parameters like filter width, filter height, stride width, stride height, and the number of filters; 2) for pooling, batchnorm, local contrast normalization, skip connections, and branching layers; 3) for a recurrent cell; and 4) for a feedforward layer, etc.
A state is implicit, and can be regarded as the partial network architecture yielded by the RNN.
With the parameters of the RNN, a neural network can be sampled, 
and it can be trained to convergence to obtain an accuracy on a validation data set, which serves as the reward, based on which the parameters of the RNN will be updated with REINFORCE, a policy gradient RL algorithm.
To reduce variance for policy gradient estimation, a baseline is included by taking an exponential moving average of the previous validation accuracies. 

Experiments are conducted on CIFAR-10 for image classification and the Penn Treebank dataset for language modelling.

It is computationally expensive to search for good neural network architectures over a large dataset like ImageNet.
It is thus desirable to transfer a good architecture searched over a small dataset like CIFAR-10 to ImageNet.

An approach is to search for the best generic convolutional layer or cell structure consisting of convolutional filters, nonlinearities, and connections as the motif to repeat multiple times to build convolutional networks.
Two types of convolutional layers are designed for image classification: 
1) a Normal Cell returns a feature map of the same dimension and
2) a Reduction Cell returns a feature map with the height and width halved.
An RNN is used to search for the structures of a Normal Cell and a Reduction Cell with RL.
The RNN predicts each cell as a group of multiple blocks, and each block has 5 prediction steps:
in Step 1 and 2, select two hidden states respectively from two outputs from lower cells or the input image, or two hidden states from previous blocks;
in Step 3 and 4, select operations to apply to the hidden states in Step 1 and 2, respectively; 
in Step 5, select a method, element-wise addition or concatenation, to combine outputs of Step 3 and 4. 

In Step 3 and 4, the RNN chooses operations from the following prevalent ones:
identity, 
1 $\times$ 3 then 3 $\times$ 1 convolution, 
1 $\times$ 7 then 7 $\times$ 1 convolution, 
3 $\times$ 3 dilated convolution,
3 $\times$ 3 average pooling,
3 $\times$ 3 max pooling,
5 $\times$ 5 max pooling,
7 $\times$ 7 max pooling,
1 $\times$ 1 convolution,
3 $\times$ 3 convolution,
3 $\times$ 3 depth-wise-separable convolution,
5 $\times$ 5 depth-wise-separable convolution, and
7 $\times$ 7 depth-wise-separable convolution.

The overall architectures of convolutional networks are predetermined manually, 
with multiple normal cells being stacked between reduction cells,
and the number of normal cells is a hyperparameter.

Experiments show that, convolutional neural networks with such transferable structures can achieve comparable or better performance on CIFAR-10 and ImageNet, as well as on mobile and embedded platforms.

\subsection{Device Placement}

The computational requirements for training and inference with neural networks are increasing recently. 
A common approach is to use a heterogeneous environment including a mixture of hardware devices like CPUs and GPUs.
How to optimize device placement, e.g., placing TensorFlow computational graphs onto hardware devices, thus becomes an issue. 
In the following, we discuss an approach with RL.  

\begin{figure}[h]
\centering
\includegraphics[width=1.0\linewidth]{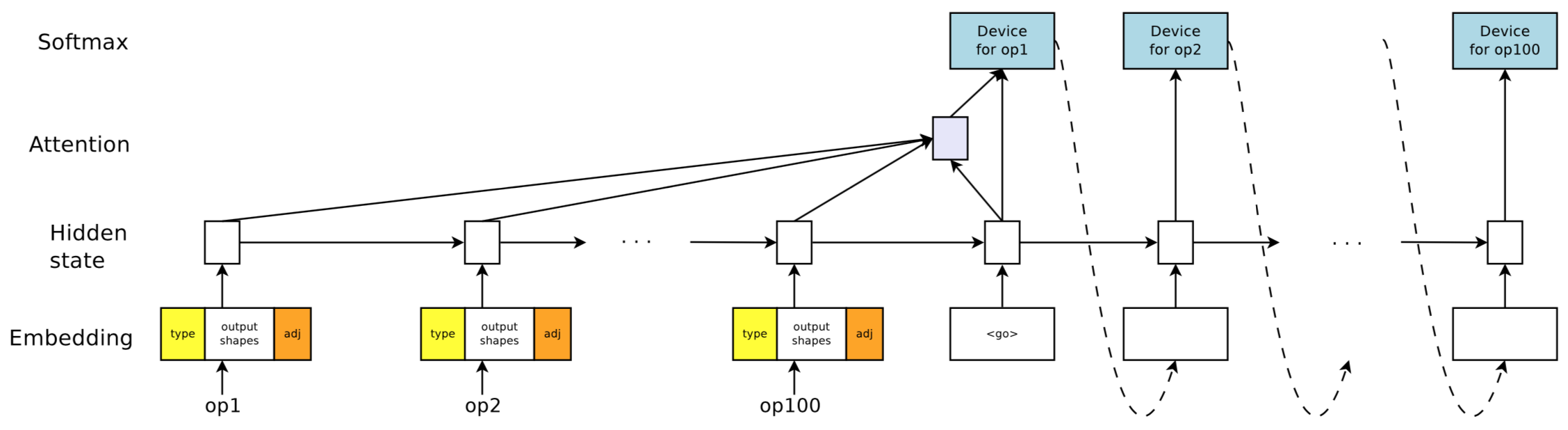}
\caption{Device placement model architecture. (from \cite{Mirhoseini2017})}
\label{DevicePlacement}
\end{figure}

A sequence-to-sequence model with long short-term memory (LSTM) and an attention mechanism are used to predict how to place operations in a TensorFlow graph onto available devices.
The sequence of operations is input to the encoder recurrent neural network (RNN), 
and is embedded by concatenating their information about type, output shape, and adjacency information.
The decoder is an attentional LSTM with the number of time steps equal to the number of operations in the computational graph.
At each step, the decoder output a device to place the corresponding operation in the encoder, 
and the embedding of the device becomes an input to the decoder in the next step.
Figure~\ref{DevicePlacement} illustrates the device placement model architecture.

With the parameters of the encoder and decoder RNNs and the attention mechanism,
we can sample a device placement and run it to check the performance of running time.
The reward signal is the square root of running time.
The parameters are tuned with an RL algorithm, in particular, REINFORCE;
and a moving average baseline is used to reduce the variance.
To improve efficiency, co-locating groups are formed either manually or automatically with the assistance of a neural network, so that operations in the same group are placed on the same device.

Experiments are conducted on ImageNet classification with Inception-V3, language modelling, and neural machine translation.

\subsection{Data Augmentation}

Data augmentation increases data amount and diversity by random augmentations.
It is desirable to automate data augmentation, rather than to do it manually, or to directly hardcode invariances into the model architecture.
In the following, we discuss how to find an optimal augmentation policy by formulating it as a discrete search problem and solving it with RL.

The search space consists of policies, each of which is composed of 5 sub-policies, 
and each sub-policy is formed by two sequential image operations with two hyperparameters for the probability of applying it and its magnitude.

We discuss image operations in the following, 
where $m$ denotes the parameter $magnitude$.
ShearX/Y, shear the image horizontally/vertically with a rate $m$;
TranslateX/Y, translate the image horizontally/vertically by the number of $m$ pixels;
Rotate, rotate the image $m$ degrees;
AutoContrast, maximize the image contrast;
Invert, invert image pixels;
Equalize, equalize the image histogram;
Solarize, invert pixels above a threshold $m$;
Posterize, reduce up to the number of $m$ bits for each pixel;
Contrast, control the image contrast; 
Color, adjust the image color balance;
Brightness, adjust the image brightness;
Sharpness, adjust the image sharpness;
Cutout, set a random square patch with side-length $m$ of pixels to grey;
Sample Pairing, add the image linearly with another image weighted by $m$, and keep the label.

\begin{figure}[h]
\centering
\includegraphics[width=0.7\linewidth]{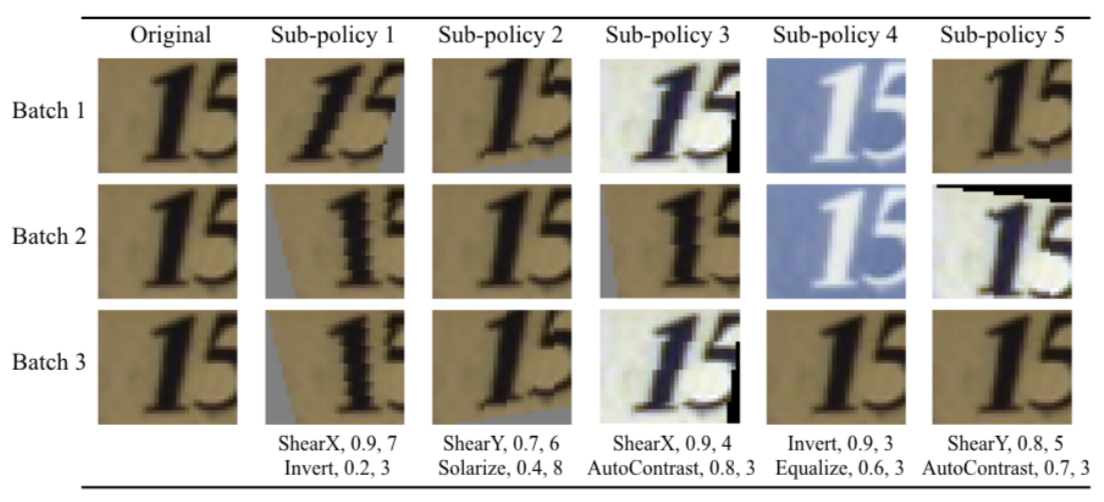}
\caption{A policy found on SVHN, and how to use it. (from \cite{Cubuk2019AutoAugment})}
\label{AutoAugment}
\end{figure}

A recurrent neural network (RNN) is used as a controller, 
which has 30 softmax predictions for 5 sub-policies, each having 2 operations, 
and each operation having operation type, magnitude, and probability.
A neural network model is trained with augmented data generated by applying the 5 sub-policies on the training dataset,
in particular, one of 5 sub-policies is selected randomly to augment an image in the mini-batch.
The accuracy of the model on the validation dataset is the reward for training the controller.
Proximal Policy Optimization (PPO) algorithm is used for training the RNN controller.
At the end, the best 5 policies, i.e., 25 sub-policies, are concatenated to form a single final policy.
Figure~\ref{AutoAugment} shows an example.
Experiments are conducted on CIFAR-10, CIFAR-100, SVHN, and ImageNet datasets.

\subsection{Cluster Scheduling}

Data processing clusters are common in data center and cloud computing.
It is critical to design efficient cluster scheduling algorithms considering workload characteristics.

Usually a directed acyclic graph (DAG) encodes job stages and dependencies.
A graph embedding with a graph neural network takes job DAGs as input and outputs three types of embeddings: 
per-node embeddings, per-job embeddings and a global embedding.
A graph embedding can encode state information like attributes of job stages and DAG dependency structure.
Nodes of DAGs carry stage attributes like the number of remaining tasks and expected task duration. 
Per-node embeddings encode information about the node and its children.
Per-job embeddings encode information about the entire job.
The global embedding encode a cluster-level summary from information of all per-job embeddings. 
Such embeddings are learned via end-to-end training.

A feature vector of a node in a job DAG is composed of: 
1) the number of tasks remaining in the stage; 
2) the average task duration; 
3) the number of executors currently working on the node;
4) the number of available executors; and 
5) whether available executors are local to the job.
%paraphrasing?

\begin{figure}[h]
\centering
\includegraphics[width=0.7\linewidth]{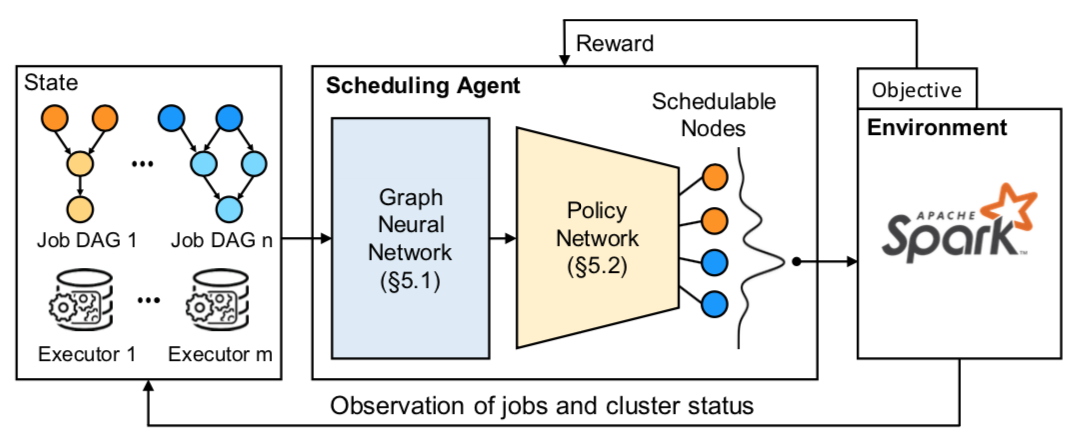}
\caption{RL framework for cluster scheduling. (from \cite{Mao2019SIGCOMM})}
\label{ClusterScheduling}
\end{figure}

Figure~\ref{ClusterScheduling} illustrates an RL framework for cluster scheduling.
When scheduling events like completion of a stage or arrival of a job occur,
the agent observes the state of the cluster, i.e., the status of the DAGs in the scheduler's queue and the executors,
uses the graph neural network to convert the state into a feature vector,
and decides on a scheduling action using the policy network.
The reward can be the average of job completion time, which is used to optimize the scheduling policy.
The reward can also be the makespan. 

\subsection{An Open Platform for Computer Systems}

Many computer systems problems are by nature sequential decision making problems,
and can be formulated as MDPs, thus RL is a solution method.
One challenge is that there are many details about computer systems. 
Park is an open platform for learning augmented computer systems, 
hiding details of computer systems to machine learning researchers.
Park provides environments for a wide range of systems problems, namely,
1) adaptive video streaming; 
2) Spark cluster job scheduling; 
3) SQL database query optimization; 
4) network congestion control; 
5) network active queue management; 
6) curcuit design;
7) Tensorflow device placement;
8) content delivery networks (CDN) memory caching;
9) account region assignment;
10) server load balancing; and,
11) switch scheduling.
Table~\ref{ComputerSystemsTable} shows their RL formulations and relevant issues, 
w.r.t. environment, type, state space, action space, reward, step time, and challenges.

\begin{table}[h]
\resizebox{\textwidth}{!}{%
\begin{tabular}{ |c|c|c|c|c|c|c| }
\hline
\bf{environment} & \bf{type} & \bf{state space} & \bf{action space} & \bf{reward} & \bf{step time} & \bf{challenges} \\ \hline \hline

\makecell{adaptive\\ video streaming} &
real/sim &
\makecell{past network throughput\\ measurements, playback\\ buffer size, portion of\\ unwatched video} &
\makecell{bitrate of the\\ next video chunk} &
\makecell{combination\\ of resolution and\\ stall time} &
\makecell{real: $\sim$3s\\ Sim: $\sim$1ms} &
\makecell{input-driven variance,\\ slow interaction time} \\ \hline

\makecell{Spark cluster\\ job scheduling} &
real/sim &
\makecell{cluster and job\\ information as features\\ attached to each node\\ of the job DAGs} &
\makecell{node to\\ schedule next} &
\makecell{runtime penalty\\ of each job} &
\makecell{real: $\sim$5s\\ Sim: $\sim$5ms} &
\makecell{input-driven variance,\\ state representation,\\ infinite horizon,\\ reality gap} \\ \hline
 
\makecell{SQL database\\ query optimization} &
real &
\makecell{query graph with\\ predicate and table\\ features on nodes,\\ join attributes on edges} &
\makecell{edge to join next} &
\makecell{cost model or\\ actual query time}&
$\sim$5s &
\makecell{state representation,\\ reality gap} \\ \hline

\makecell{network\\ congestion control} &
real &
\makecell{throughput, delay\\ and packet loss} &
\makecell{congestion window\\ and pacing rate} &
\makecell{combination of\\ throughput and delay} &
$\sim$10ms &
\makecell{sparse space for\\ exploration, safe\\ exploration, infinite\\ horizon} \\ \hline

\makecell{network active\\ queue management} &
real &
\makecell{past queuing delay,\\ enqueue/dequeue rate} &
drop rate &
\makecell{combination of\\ throughput and delay} &
$\sim$50ms &
\makecell{infinite horizon,\\ reality gap} \\ \hline
 
\makecell{Tensorflow\\ device placement} &
real/sim &
\makecell{current device placement\\ and runtime costs as\\ features attached to each\\ node of the job DAGs} &
\makecell{updated placement\\ of the current node} &
\makecell{penalty of runtime\\ and invalid placement} &
\makecell{real: $\sim$2s\\ Sim: $\sim$10ms} &
\makecell{state representation,\\ reality gap} \\ \hline

circuit design &
sim &
\makecell{circuit graph with\\ component ID, type\\ and static parameters\\ as features on the node} &
\makecell{transistor sizes,\\ capacitance and\\ resistance of\\ each node} &
\makecell{combination of\\ bandwidth, power\\ and gain} &
$\sim$2s &
\makecell{state representation,\\ sparse space for\\ exploration} \\ \hline

\makecell{CDN\\ memory caching} &
sim &
\makecell{object size, time since\\ last hit, cache occupancy} &
admit/drop &
byte hits &
$\sim$2ms &
\makecell{input-driven variance,\\ infinite horizon,\\ safe exploration} \\ \hline
 
\makecell{multi-dim database\\ indexing} &
real &
\makecell{query workload,\\ stored data points} &
\makecell{layout for data\\ organization} &
query throughput &
$\sim$30s &
\makecell{state/action\\ representation,\\ infinite horizon} \\ \hline
  
\makecell{account\\ region assignment} &
sim &
\makecell{account language,\\ region of request,\\ set of linked websites} &
\makecell{account region\\ assignment} &
\makecell{serving cost\\ in the future} &
$\sim$1ms &
\makecell{state/action\\ representation} \\ \hline  

\makecell{server load\\ balancing} &
sim &
\makecell{current load of the\\ servers and the size\\ of incoming job} &
\makecell{server ID to\\ assign the job} &
\makecell{runtime penalty\\ of each job} &
$\sim$1ms &
\makecell{input-driven variance,\\ infinite horizon,\\ safe exploration} \\ \hline

switch scheduling &
sim &
\makecell{queue occupancy for\\ input-output port pairs} &
\makecell{bijection mapping\\ from input ports\\ to output ports} &
\makecell{penalty of remaining\\ packets in the queue} &
$\sim$1ms &
action representation \\ \hline
 
\end{tabular}
}
\caption{Overview of the computer system environments supported by Park platform. (reproduced from \cite{Mao2019RL4RealLife})}
\label{ComputerSystemsTable}
\end{table}

These problems are along three dimensions, in particular,
a) global vs. distributed control;
b) fast control loop vs. planning; and,
c) real system vs. simulation.
Some environments use real systems in the backend. 
For the other environments, simulators are provided with well-understood dynamics.

The authors also discuss challenges that off-the-shelf RL algorithms may not work,
particularly, sparse meaningful space for exploration and representation for state-action space,
and input-driven variance, infinite horizon, and reality gap for decision process.

\begin{figure}[h]
\centering
\includegraphics[width=0.7\linewidth]{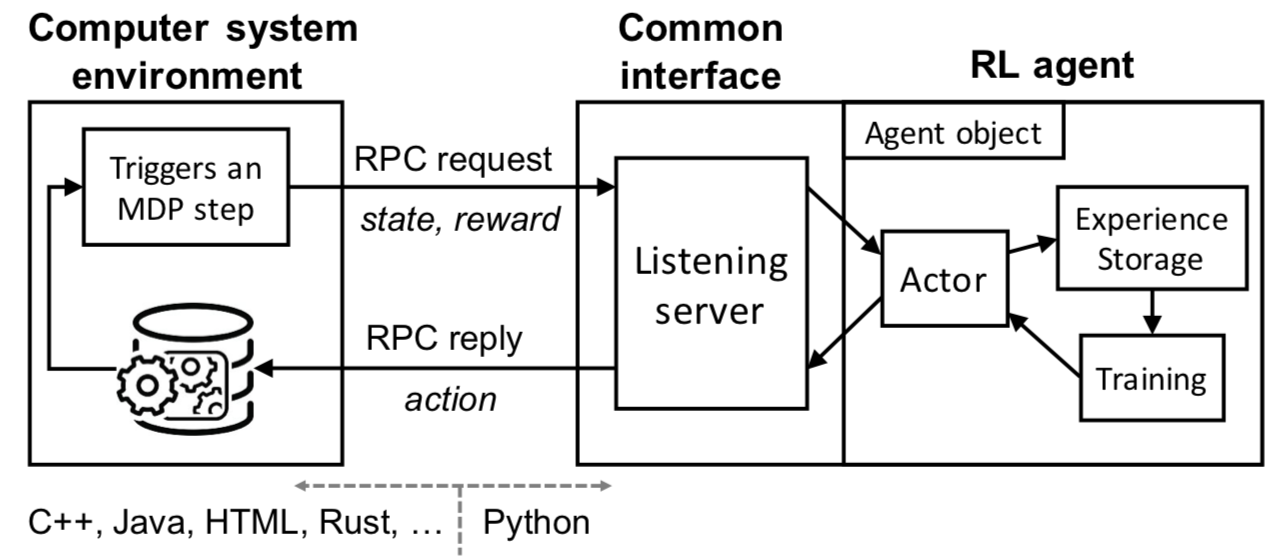}
\caption{Park architects an RL-as-a-service design paradigm. (from \cite{Mao2019RL4RealLife})}
\label{ComputerSystems}
\end{figure}

As shown in Figure~\ref{ComputerSystems}, an RL agent connects with the computer system through a request/response interface, which make the system complexity transparent.
An RL agent can interacts with a real system or a simulator.

\subsection{NP-Hard Problems}

NP-hard problems, e.g., minimum vertex cover, maximum cut, travelling salesman problems (TSP), and graph coloring, have many applications, like planning, scheduling, logistics, compilers, and design automation.
However, we do not know polynomial time algorithms for the hard ones among them, like those mentioned above.
Note the hardness is in the sense of worst case complexity, and in practice, even very large problems can be solved, approximately or heuristically.
It is therefore desirable to design efficient and effective algorithms to solve such large-scale problems.

NP-hard problems may be tackled with exact, approximate, or heuristic algorithms.\footnote{Here, heuristic algorithms usually yield inexact solutions. Heuristic search, a branch in AI, like A* and IDA*, guarantees exact solutions if solved. Heuristic search is close to branch-and-bound.}
Exact algorithms may use enumeration or branch-and-bound, may be formulated as integer programming, and are usually prohibitive for large instances.
Approximate algorithms have polynomial time complexity, with or even without sub-optimal performance guarantees.
Heuristic algorithms are usually fast, yet without performance guarantee, and usually need considerable manual design.
We discuss how to learn good heuristics for several graph combinatorial optimization problems with RL.

In the following we give brief definitions of several problems. 
In minimum vertex cover, we find a subset of nodes on a graph to cover all edges, and the number of nodes is minimal.
In maximum cut, we find a subset of nodes on a graph to maximize the weight of the cut edges forming with nodes not in the subset.
In TSP, we find a tour to visit all the nodes only once and to minimize the total weight.
In coloring, we assign the minimal number of colors to nodes so that adjacent nodes are of different colors.

\subsubsection*{Minimum Vertex Cover, Maximum Cut, and TSP}

We first discuss RL formulations for addressing the problems of 
minimum vertex cover, maximum cut, and TSP.
Nodes in a graph are represented with embeddings to allow algorithmic flexibility with diverse graphs.
A state is a sequence of actions to choose nodes on a graph.
An action is to choose a node not part of the current partial solution.
Both a state and an action are represented by the node embeddings.
The reward is the change in the cost function after taking an action in a state.
The policy specifies which node to choose at the current state.

For minimum vertex cover, 
a state is the subset of nodes in the current partial solution,
an action is to add a node to the partial solution,
the reward is -1,
and it terminates when all edges are covered.
For maximum cut, 
a state is the subset of nodes in the current partial solution,
an action is to add a node to the partial solution,
the reward is the change in cut weight,
and it terminates when the cut weight stops improving.
For TSP, 
a state is the partial tour,
an action is to add a node to the partial tour,
the reward is the change in tour cost,
and it terminates when the tour contains all nodes.

A combination of multi-step Q-learning and fitted Q-iteration are used to learn the action value Q function, which can derive an optimal policy directly.
Experiments are conducted on generated graphs with number of nodes from 50 to 1200. 

\subsubsection*{AlphaGo Zero}

We give a brief review of AlphaGo Zero in the following.
It is followed to learn an optimal policy for graph coloring as we discussed later.

AlphaGo Zero 1) learns from random play, with self-play RL, without human data or supervision; 
(2) does not need manual feature engineering; 
(3) uses a single neural network to represent both policy and value; 
and (4) utilizes the neural network for state evaluation and action sampling for MCTS, and it does not perform Monte Carlo rollouts.
 
AlphaGo Zero deploys several recent achievements in neural networks: ResNets, batch normalization, and rectifier nonlinearities.
AlphaGo Zero has three main components in its self-play training pipeline executed in parallel asynchronously: 
(1) optimize neural network weights from recent self-play data continually; 
(2) evaluate players continually; 
and (3) use the strongest player to generate new self-play data.
When AlphaGo Zero playing a game against an opponent, MCTS searches from the current state, with the trained neural network weights, to generate move probabilities, and then selects a move.

AlphaGo Zero follows a generalized policy iteration (GPI) scheme, incorporating Monte Carlo tree search (MCTS) inside the training loop to perform both policy evaluation and policy improvement. 
MCTS may be regarded as a policy improvement operator,
outputing action probabilities stronger than raw probabilities of the neural network. 
Self-play with search may be regarded as a policy evaluation operator, 
using MCTS to select actions, and game winners as samples of value function. 
Then the policy iteration procedure updates the neural network's weights to match action probabilities and value respectively more closely with the improved search probabilities and self-play winner, and conducts self-play with updated neural network weights in the next iteration to make the search stronger.

\begin{figure}[h]
\centering
\includegraphics[width=0.7\linewidth]{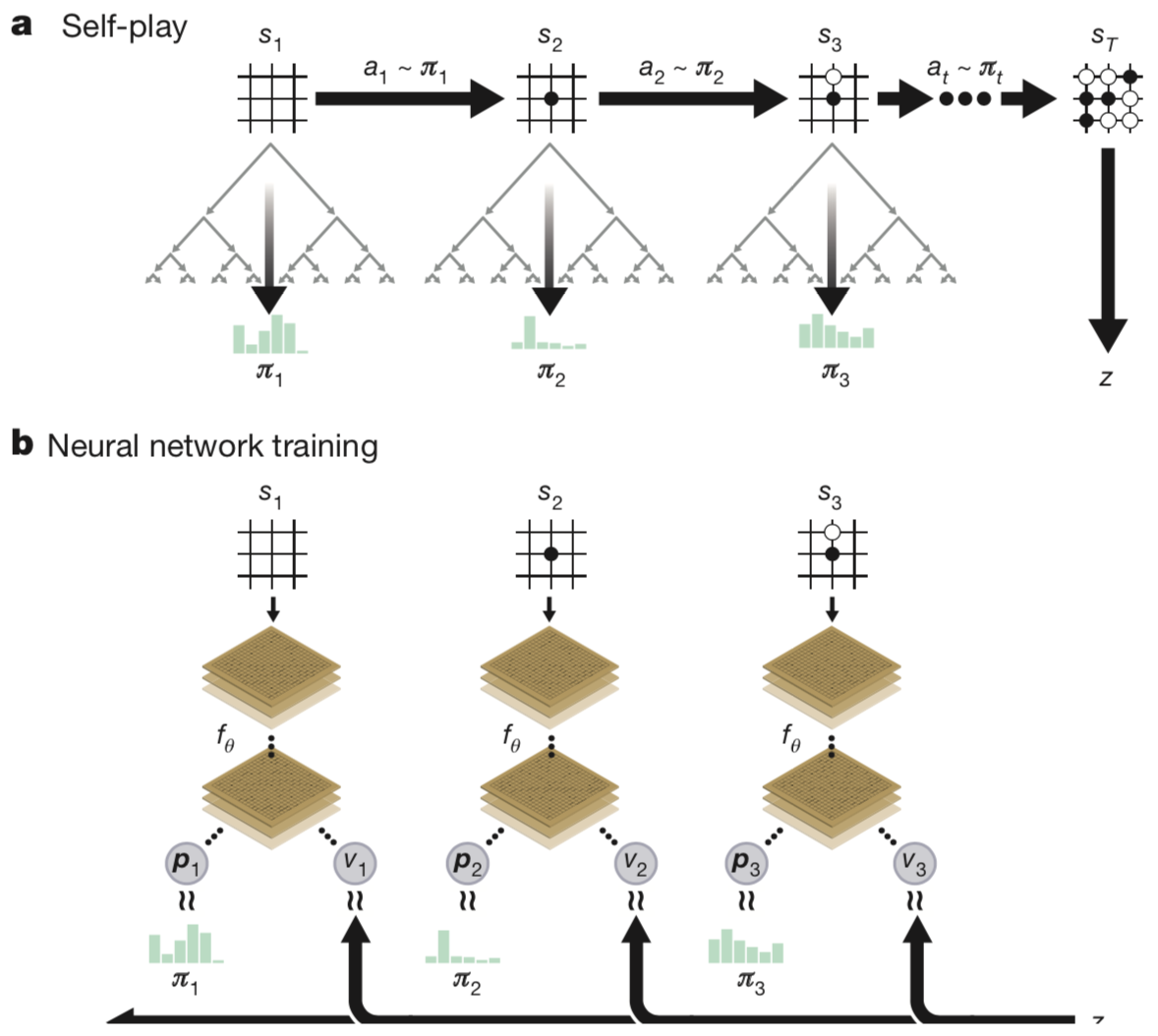}
\caption{Self-play RL in AlphaGo Zero. (from \cite{Silver-AlphaGo-2017})}
\label{AlphaGoZeroSelfPlay}
\end{figure}

As shown in Figure~\ref{AlphaGoZeroSelfPlay}, in Step \textbf{a} with self play, AlphaGo Zero plays a program against itself;
in Step \textbf{b},  AlphaGo Zero trains the neural network with data collected from self play.

\begin{figure}[h]
\centering
\includegraphics[width=1.0\linewidth]{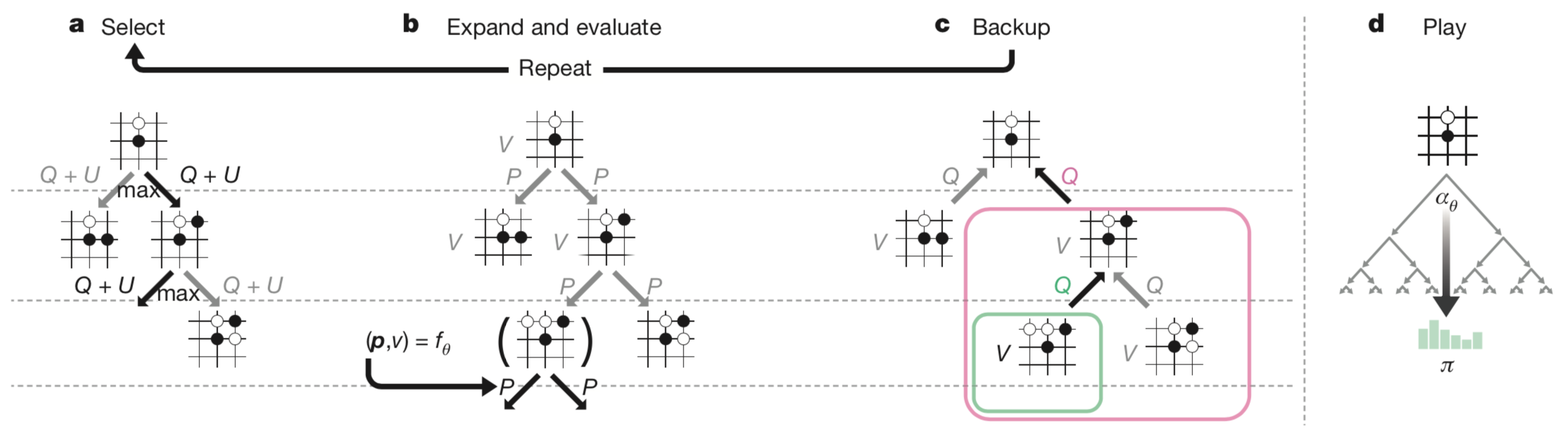}
\caption{MCTS in AlphaGo Zero. (from \cite{Silver-AlphaGo-2017})}
\label{AlphaGoZeroMCTS}
\end{figure}

As shown in Figure~\ref{AlphaGoZeroMCTS}, 
in Step \textbf{a}, MCTS selects a move which most likely leads to a good result;
in Step \textbf{b}, MCTS expands the leaf node and evaluates the associated position by the neural network;
in Step \textbf{c}, MCTS backups information in the subtree traversed; and,
in Step \textbf{d}, MCTS plays the game with the updated search probabilities.

\subsubsection*{Graph Coloring}

Next we discuss an RL formulation for graph coloring.
A state represents the current partial assignment of colors to nodes.
An action is to assign a valid color to the next node.
One way to define a reward is the negative total number of colors assigned so far.
A policy maps a state to an action.
To treat the graph coloring as a zero sum game, and to deploy self-play, 
an alternative way to define a reward is similar to computer Go, with win, lose and tie,
for the cases that the policy has fewer, more, or the same number of colors,
comparing with the current best policy. 

For graph coloring, there are graph context, problem context, and colors context.
Graph context contains embeddings for 
the number of nodes in the graph, 
the number of assigned colors,
and the number of already colored nodes.
Problem context contains embeddings for
nodes just been colored and nodes scheduled for coloring next.
Possible colors context contains embeddings for
nodes colored with possible colors.

A probability network computes the probability to assign a valid color to the current node,
taking inputs of graph context, problem context, and context for each possible color.
A value network computes the expected outcome of the coloring problem, i.e., win, lose, or tie, 
using problem context and colors context.

Comparing with computer Go, in graph coloring, we need to deal with diverse graphs, and problem sizes can be much larger. 
In experiments, the number of nodes varies from 32 to 10 million, and may require up to 300 GPUs for training.
After training, we may distill the best heuristics into models as production tools.

\subsection{Annotated Bibliography}

The discussion about neural architecture search is based on \cite{Zoph2017}.
The discussion about transferable architecture search is based on \cite{Zoph2017Transfer}.
The discussion about  device placement is based on \cite{Mirhoseini2017}.
The discussion about  data augmentation is based on \cite{Cubuk2019AutoAugment}.
These are part of the effort of AutoML.
See an overview of Google's work on AutoML and future directions, video at \url{https://slideslive.com/38917182}.

The discussion about cluster scheduling is based on \cite{Mao2019SIGCOMM}.
The discussion about Park, an open platform for learning augmented computer systems, is based on~\cite{Mao2019RL4RealLife}.
See SysML Conference at \url{https://www.sysml.cc}.

The discussion about AlphaGo Zero is based on \cite{Silver-AlphaGo-2017}.
The discussion about minimum vertex cover, maximum cut, and TSP is based on \cite{Dai2017graph}.
The discussion about graph coloring is based on \cite{HuangJiayi2019}.
\cite{Kool2019ICLR} study an attention approach for vehicle routing problems.
\cite{Segler2018} study chemical syntheses with AlphaGo techniques.

\clearpage

\section{Energy}
\label{energy}

Our mankind is facing issues of sustainable and green energy.
It is critical to consume energy efficiently. 
We discuss data center cooling and smart grid in the following.

\subsection{Data Center Cooling}

Data centers are widespread in the IT era, esp. after the prevalence of big data and AI.
Cooling is essential for data center infrastructure  to lower the high temperature and to reduce the amount of heat 
in order to improve the performance and to mitigate the potential of equipment damage.
There are difficulties like unexpected events, safety constraints, limited data, and potentially expensive failures.
RL is one approach to data center cooling.

We discuss how to control fan speeds and water flow in air handling units (AHUs) to regulate the airflow and temperature inside server floors, using model-predictive control (MPC).
In MPC, the controller (agent) learns a linear model of the data center dynamics with random, safe exploration, with little or no prior knowledge.
It then optimizes the cost (reward) of a trajectory based on the predicted model, 
and generates actions at each step, to mitigate the effect of model error and unexpected disturbances, at the cost of extra computation.

\begin{figure}[h]
\centering
\includegraphics[width=0.8\linewidth]{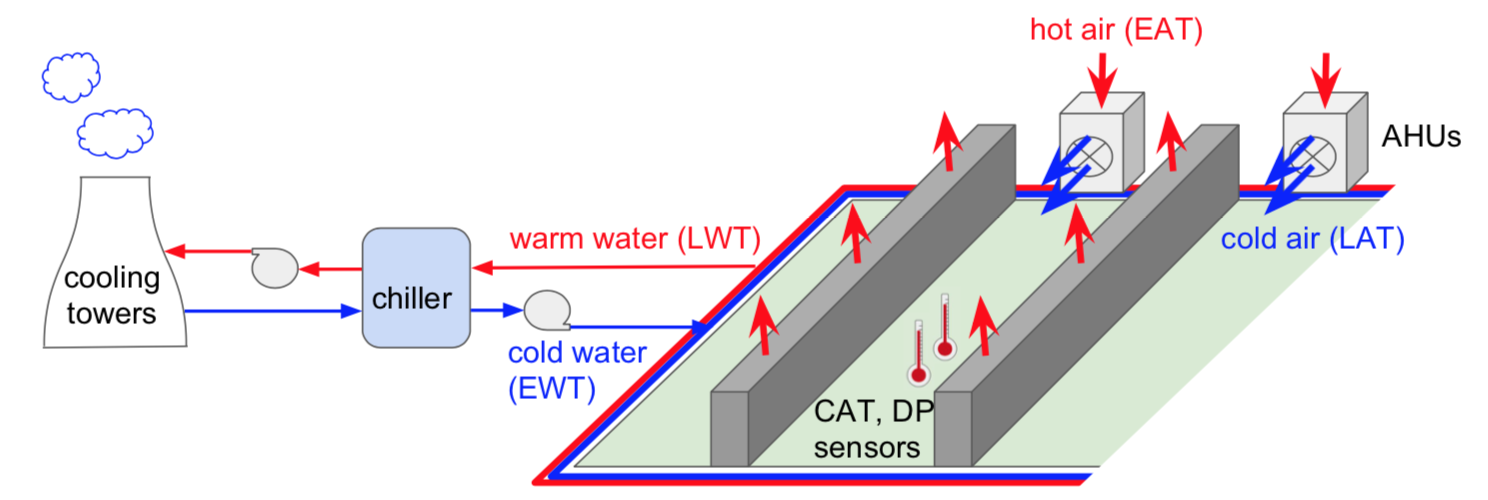}
\caption{Data center cooling loop. (from \cite{Lazic2018NIPS})}
\label{DataCenterCooling}
\end{figure}

The controls or actions are variables to manipulate, including fan speed to control air flow and valve opening to regulate the amount of water.
States refer to the process variables to predict and regulate, including 
differential air pressure (DP),
cold-aisle temperature (CAT),
entering air temperature (EAT) to each AHU,
leaving air temperature (LAT) from each AHU.
There are also system disturbances, referring to events or conditions out of manipulation or control, including,
server power usage to surrogate the amount of generated heat,
and entering water temperature (EWT) of the chilled water at each AHU. 
Figure~\ref{DataCenterCooling} illustrates the data center cooling loop.
 
The method is compared with a local proportional integral derivative (PID) controller and a certainty-equivalent controller, 
which updates parameters of the dynamics model assuming the estimated model were accurate. 
Experiments show that the method can achieve data center cooling in a large-scale commercial system in a safe, effective, and cost-efficient way.

\subsection{Smart Grid}

A smart grid may have many new components, namely,
1) advanced metering infrastructure (AMI), demand response and curtailable loads;
2) flexible power electronics;
3) photovoltaics and solar heating;
4) recharging electronic vehicle;
5) microgrid;
6) energy storage;
7) distributed generation;
8) storm management;
9) massive solar thermal and wind generation facilities; and
10) nanotechnologies.

\begin{figure}[h]
\centering
\includegraphics[width=0.6\linewidth]{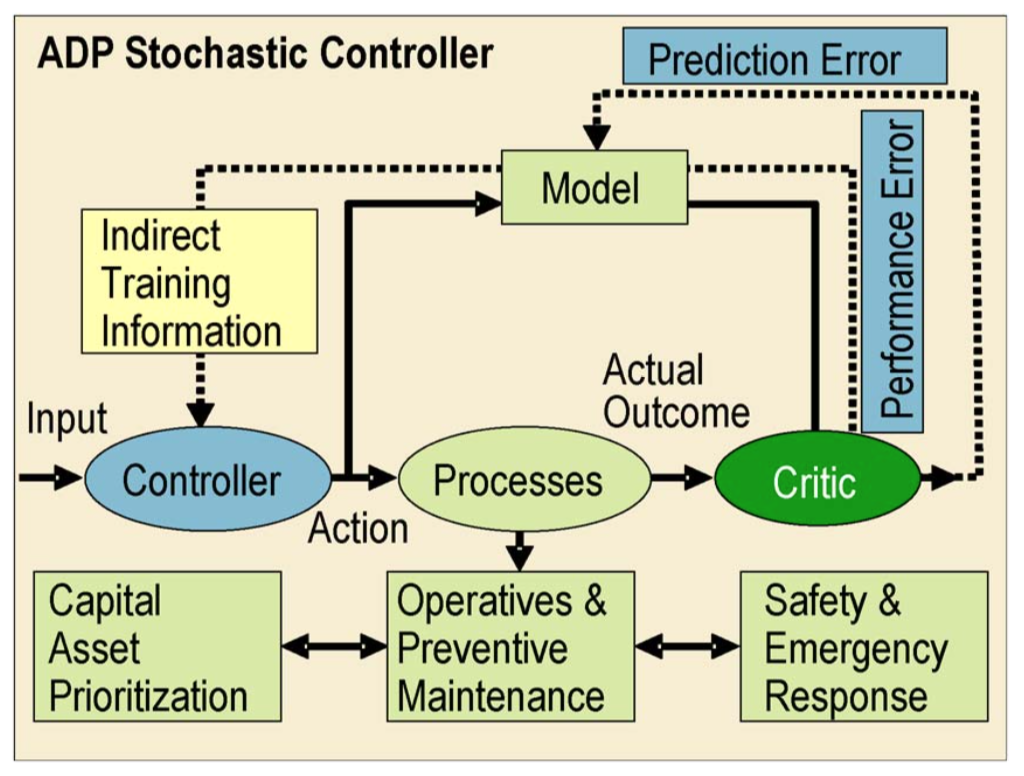}
\caption{The feedback loops for an adaptive stochastic controller for the smart grid. (from \cite{Anderson2011})}
\label{SmartGrid}
\end{figure}

As shown in Figure~\ref{SmartGrid},  the feedback loops for an adaptive stochastic controller for the smart grid 
takes data from distributed sources as input and 
optimizes capital asset prioritization, 
operative \& preventive maintenance tasks, 
and safety \& emergency responses simultaneously.

We discuss an approximate dynamic programming or RL formulation for the smart grid.
A state variable consists of  a physical state, an information state and a belief state.
A physical state may include amount of energy in a battery, status of a diesel generator (on/off), or other physical dimensions of the system.
An information state may include current and historical demand, energy availability from wind/solar, etc., and electricity prices.
A belief refers to uncertainty in the state of the system, like probability distributions of demand, the reliability of the network, or best prices, etc.
A state also includes the exogenous information, which refers to
possible changes in demand, and/or supply of energy, 
as well as network contingencies and emergency failures.
Decisions,  or actions or controls,
include whether to charge or discharge the battery, draw power from the grid or pump power into it, use backup generation, etc.
The transition function
consists of a set of equations describing the evolving of the system over time,
and, given the state and the decision, determines the next state.
The objective function governs how we make decisions and evaluates the performance of the designed policies.
The goal is to find optimal decisions to minimize some cost or maximize some reward function, 
and at the same time, maintaining the risk or safety level.   

\subsection{Annotated Bibliography}

The discussion about data center cooling is based on \cite{Lazic2018NIPS}.
The discussion about smart grid is based on \cite{Anderson2011}.

\clearpage

\section{Finance}
\label{finance}

Many problems in finance and economics are decision problems with sequential nature, e.g., asset pricing, portfolio optimization, risk management, and RL is a natural solution method. In the following, we discuss case studies for option pricing and order book execution.

\subsection{Option Pricing}

Options are important financial instruments, and are ubiquitous in all kinds of financial markets, e.g., equity, fixed income, convertible, commodity, and mortgage.
A call option gives the holder the right, not the obligation, to buy an underlying asset, e.g., a share of a stock, by a certain date, i.e., the maturity date, for a certain price, i.e., the strike price.
Similarly, a put option gives the holder the right to sell an underlying asset by the maturity date for the strike price.
A European style option can be exercised only at the maturity date.
An American style option has the flexibility to be exercised at any time up to the maturity date.
A challenging problem is to determine the fair price of an option, i.e., option pricing.  
  
To formulate option pricing as an RL problem, we need to define states, actions, and rewards.  
Here we discuss stock option, and the discussion is general for options with other types of underlying assets.
A stock option is contingent on multiple sources of uncertainty, each of which is part of the state, thus the state may be high dimensional.
The stock price is a state variable. 
Financial and economic factors, e.g., interest rates, can be part of the state.
An option's price is related to the time, thus time can be part of the state too.

For options, there  are two possible actions/strategies: exercise (stop) and continue.  
When exercising the option, the state action value function is the intrinsic value of the option, which can be calculated exactly. 
This is the reward for taking the action of exercise.
When taking an action of continue, there is no immediate reward, or the reward is 0.

A value-based RL formulation to the option pricing problem then boils down to learning the state action value function for continuation. 
There are several methods: 
1) a least square Monte Carlo (LSM) method to calculate the expected payoff from continuation backwards from the last time step to the first time step; 
2) an approximate value iteration method; 
and 3) a method based on least square policy iteration (LSPI). These methods also provide empirical and theoretical results.

\subsection{Order Book Execution}

Order book execution is a fundamental trading problem.
In limit order markets, we specify not only the desired volumes, but also the desired prices.
We discuss a simple form, in which, we need to optimize an execution strategy to trade a certain number of shares of a stock within a certain time horizon.

A state can be composed of private variables as well as market variable.
Private variables include the elapsed time and the remaining inventory.
Market variables include bid-ask spread, bid-ask volume imbalance, signed transaction volume, and immediate market order cost.
Bid-ask spread is the difference between bid and ask prices in the current order books. 
Bid-ask volume imbalance is the difference of the numbers of shares at the bid and ask, respectively, in the current order books. 
Signed transaction volume is the difference of the number of shares bought and sold, respectively, in the last 15 seconds.
Immediate market order cost is the cost to trade the remaining shares immediately with a market order.

An action is to reposition the limit order for the remaining volume relative to the current ask or bid, so that at any moment, there is a single outstanding order. 
The cost is the expenditure from a (partial) execution of an order.  
For normalizing price differences across stocks, performance is measured by the implementation shortfall, which is the difference between the average price paid and the midpoint of the ask bid spread at the start of the trading period.
The lower the implementation shortfall, the better.

Experiments are conducted on 1.5 years of millisecond time-scale limit order data for Amazon, Nvidia and Qualcomm stocks from Nasdaq.
The results confirm that, 1) for shorter trading periods and higher target volumes, trading cost are higher overall;
and 2) the execution policy becomes more aggressive, 
2.1) when the ask bid spread is large,
and 2.2) when the immediate cost of trading the remaining shares is low, a bargaining is available.

\subsection{Annotated Bibliography}

\citet{Hull06} provides an introduction to options and other financial derivatives.
Simulation based methods like RL or approximate dynamic programming are flexible to deal with complex, high dimensional scenarios. 

\cite{Longstaff01} study the least squares Monte Carlo (LSM) method.
It is a standard approach in the finance literature for pricing American options, following approximate dynamic programming. 
\cite{Tsitsiklis01} propose an approximate value iteration approach. 
\cite{Li2009Option} adapt least square policy iteration (LSPI) in~\cite{Lagoudakis03}  to the problem of American option pricing.

The discussion of order book execution is based on \cite{Kearns2013}.

In addition, \cite{Bacoyannis2018} discuss idiosyncrasies and challenges of data driven learning in electronic trading. 

%\citet{Moody01} investigate learning to trade via direct reinforcement.

\clearpage

\section{Healthcare}
\label{healthcare}

There are many opportunities and challenges in healthcare for AI, in particular, for RL.
When applying RL to healthcare, we need to consider several guidelines.
First, it is desirable for an RL algorithm to have access to all information that impact decision making.
It is important for an RL algorithm to consider the information accessible to clinicians. 
Second, the effective sample size is related to the closeness between the learned policies and clinician policies; 
the closer the policies, the larger the effective sample size.
Third, it is essential to introspect the learned policies to achieve sound behaviours.

In the following, we discuss dynamic treatment strategies and medical image report generation.

\subsection{Dynamic Treatment Strategies}

Dynamic treatment regimes (DTRs) or adaptive treatment strategies are sequential decision making problems. 
We discuss an approach to optimal treatment strategies for sepsis in intensive care.

\begin{figure}[h]
\centering
\includegraphics[width=1.0\linewidth]{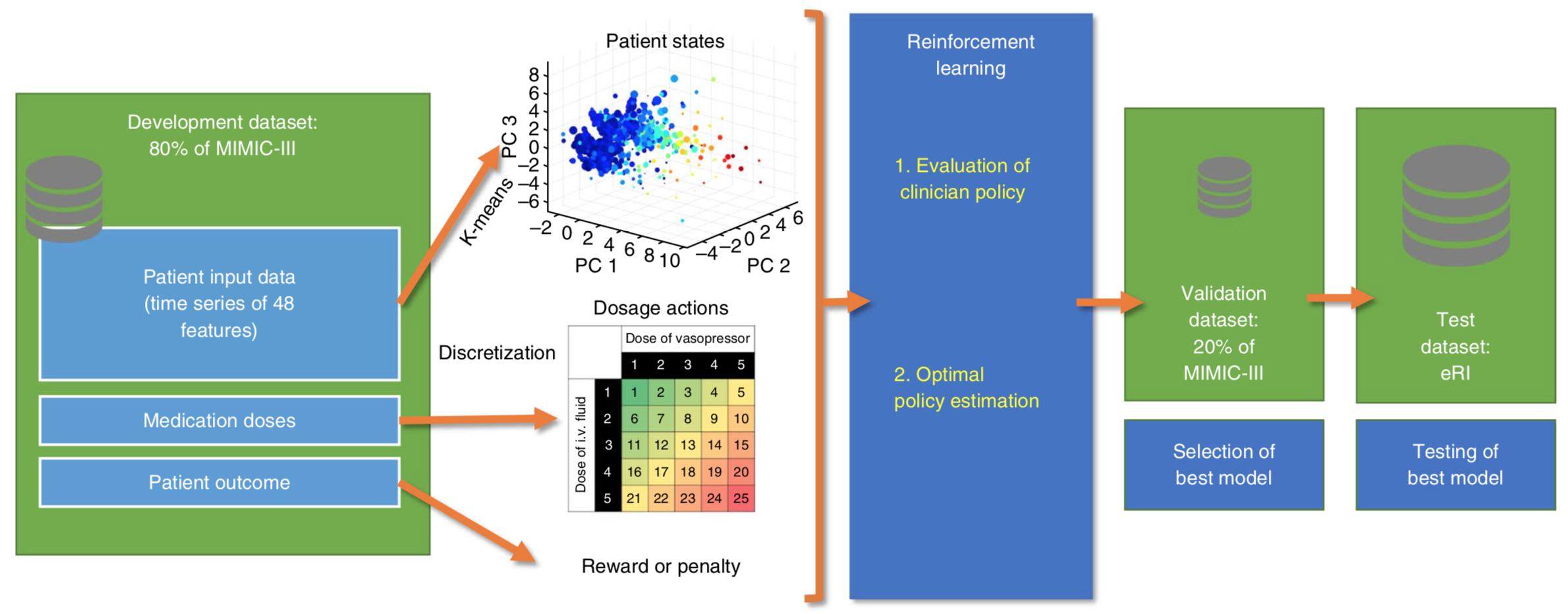}
\caption{Data flow for sepsis treatment. (from \cite{Komorowski2018})}
\label{AIClinician}
\end{figure}

A state is constructed from the multidimensional discrete time series 
composed of 48 variables about demographics, vital signs, premorbid status, laboratory values, and intravenous fluids and vasopressors received as treatments.
Clustering is used to define the state space so that patients in the same cluster are similar w.r.t. the observable properties.
An action, or a medical treatment, is defined by the total volume of intravenous fluids and maximum dose of vasopressors over each 4 hour period. 
The dose of each treatment is divided into 5 possible choices, resulting in 25 discrete actions when combining the two treatments. 
A reward and a penalty is associated with survival and death, respectively, to optimize patient mortality.
Two datasets are involved: the Medical Information Mart for Intensive Care version III (MIMIC-III) and eICU Research Institute Database (eRI).
A transition model is estimated with 80\% of the MIMIC-III dataset.
An optimal policy is learned with policy iteration, to find a policy to maximize the expected 90-day survival of patients in the MIMIC-III cohort.
20\% of the MIMIC-III dataset is used for validation of the learned models.
After an optimal model is selected, the eRI dataset is used for testing.
Since the optimal policy is usually not the same as clinician policies, 
and we need to evaluate the policy with the data generated by those clinician policies,
we resort to off-policy policy evaluation.
Figure~\ref{AIClinician} illustrates the data flow.

Experiments show that the policy learned by RL has larger value, or lower mortality, than those from human clinicians,
and that the patients have the lowest mortality when they receive treatments similar to those recommended by the policy learned by RL.

\subsection{Medical Image Report Generation}

It is challenging to generate reports to obtain a long narrative including multiple sentences or paragraphs,
with consistent topics and a reasonable logic, 
using task-specific vocabulary and content coverage.
Medical image report generation needs to follow critical protocols and to use medical terms correctly.
A medical report includes:
1) a findings section to describe both normal and abnormal observations, 
covering important aspects and potential diseases,
e.g., for chest x-ray images, aspects like heart size, lung opacity, bone structure, and abnormal phenomena with lungs, aortic and hilum, and potential diseases like effusion, consolidation and pneumothroax;
2) an impression or conclusion sentence to indicate the most outstanding observation or conclusion;
and 3) comparison and indication sections to list secondary information.

Most findings in medical reports are normal,
and reports about normal observations are dominated by a certain types of sentences,
thus choosing one of these templates has high potential to generate a good sentence for an observation.
In contrast, abnormal findings are rare, diverse, yet important.

\begin{figure}[h]
\centering
\includegraphics[width=1.0\linewidth]{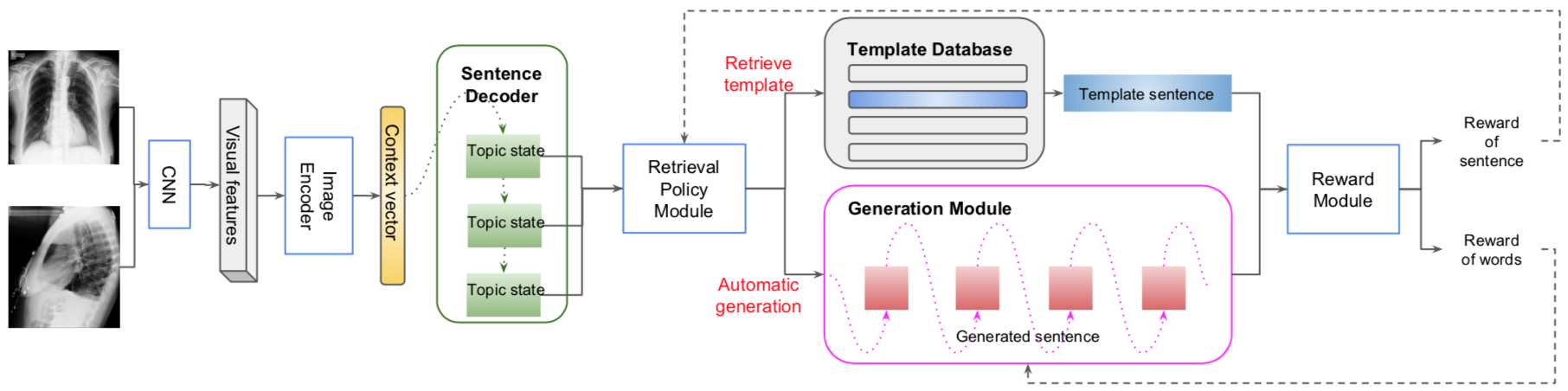}
\caption{Hybrid retrieval-generation RL agent. (from \cite{LiYuan2018NIPS})}
\label{MedicalImageReport}
\end{figure}

One way to generate medical image report is to follow a hybrid retrieval-generation approach trained by RL,
to integrate human prior knowledge and neural networks.
A convolutional neutral network (CNN) extract visual features of a set of images of a sample,
and an image encoder transforms the visual features into a context vector,  
then a sentence decoder generates latent topics recurrently.
Based on a latent topic, a retrieval policy module decides between two methods to generate sentences: 
1) a generation approach to generate sentences by an automatic generation module, and
2) a template approach to retrieve an existing template from the template database.
The retrieval policy module makes discrete decisions among an automatic generation approach and multiple templates.
The generative module also makes discrete decisions to generate a sentence from a vocabulary based on the latent topic and the image context vector.
The template database is constructed from human prior knowledge based on human medical reports,
with templates like \enquote{No pneumothorax or pleural effusion} and \enquote{Lungs are clear}.
The retrieval module and the generation module are trained jointly using REINFORCE with sentence-level and word-level rewards, respectively.
See Figure\ref{MedicalImageReport} for an illustration.

Experiments are conducted on two medical image report datasets, 
against various evaluation metrics, including automatic metrics, human evaluation, and medical terminology precision detection.

%\subsection{Mobile Health???}

\subsection{Annotated Bibliography}

The discussion about guidelines for RL in healthcare is based on \citet{Gottesman2019}. 

The discussion about individualized sepsis treatment using RL is based on \cite{Komorowski2018} and comments on it in \citet{Saria2018}.
\cite{Komorowski2018} deploy the off-policy evaluation technique in \cite{Thomas2015}.
There are recent work on off-policy evaluation, e.g., \cite{Jiang2016Doubly} and \cite{LiuYao2018NIPS}.
\cite{LiuYao2018NIPS} show a case on an HIV treatment simulation study.

See \citet{Chakraborty2014} for a recent survey about dynamic treatment regimes.

The discussion about medical image report generation is based on \citet{LiYuan2018NIPS}, 
which proposes a hybrid retrieval-generation RL approach.

\clearpage

\section{Robotics}
\label{robotics}

Robotics is a traditional application area for RL 
and can be applied to a large range of areas, e.g., manufacture, supply chain, healthcare, etc.

Robotics pose challenges to RL,
including dimensionality, real-world examples, 
under-modeling (models not capturing all details of system dynamics), model uncertainty,
reward design, and goal specification. 

RL provides the following tractable approaches to robotics:   
1) representation, including state-action discretization, value function approximation, and pre-structured policies;
2) prior knowledge, including demonstration, task structuring, and directing exploration; 
and 3) models, including mental rehearsal, which deals with simulation bias, real-world stochasticity, and optimization efficiency with simulation samples, and approaches for learned forward models.

RL algorithms are usually sample intensive and
exploration may generate policies risky to the robot and/or the environment.
It is thus easier to train a robot in simulation than in reality.  
However, there is usually a gap between simulation and reality  w.r.t. dynamics and perception,
so that a simulator usually can not precisely reflect the reality. 
We can attempt to reduce the reality gap to achieve better sim-to-real transfer by
1) improving the simulator either analytically or using data, which is usually non-trivial;
and 2) designing a policy  robust to system properties, assuming imperfect simulation.
One way to achieve robustness is by randomization, 
e.g, using a stochastic policy, randomizing the dynamics, adding noise to the observations, and perturbing the system with random disturbances.

For some RL problems, reward functions may not be available or hard to specify. 
In imitation learning, an agent learns to perform a task from expert demonstrations without reinforcement signals. 
Behavioral cloning and inverse RL are two main approaches for imitation learning. 
Behavioral cloning, or learning from demonstration, learns a policy from expert trajectories, without learning the reward function. 
It is supervised learning sometimes. 
Inverse RL learns a reward function first.

In the following, we introduce several work using sim-to-real and imitation learning for robotics.

\subsection{Dexterous Robot}

Robots are commonly in use for structured settings like pipelines in factories.
However, it is desirable to design manipulators to perform tasks, e.g. moving objects, for unstructured environments like in homes and at hospitals. 
Multi-fingered hands are versatile manipulators.

In the following, we discuss in-hand object reorientation.
An object, e.g., a block or an octagonal prism, is placed on the palm of a humanoid robot,
and we aim to reorient the object to a target configuration in hand. 
We use a humanoid robotic hand Shadow Dexterous Hand.

Observations are fingertip positions, object position, object orientation, target orientation, relative target orientation, hand joints angles, hand joints velocities, object velocity, and object angular velocity.
Actions are desired joints angles relative to current ones.
At each time step, there is a difference of rotation angles between the desired and the current object orientation.
The reward is defined for each transition as the difference before the transition minus the difference after the transition.
In addition, a reward of 5 is given for achieving a goal, and a penalty of -20 is given for dropping the object.
 
\begin{figure}[h]
\centering
\includegraphics[width=0.8\linewidth]{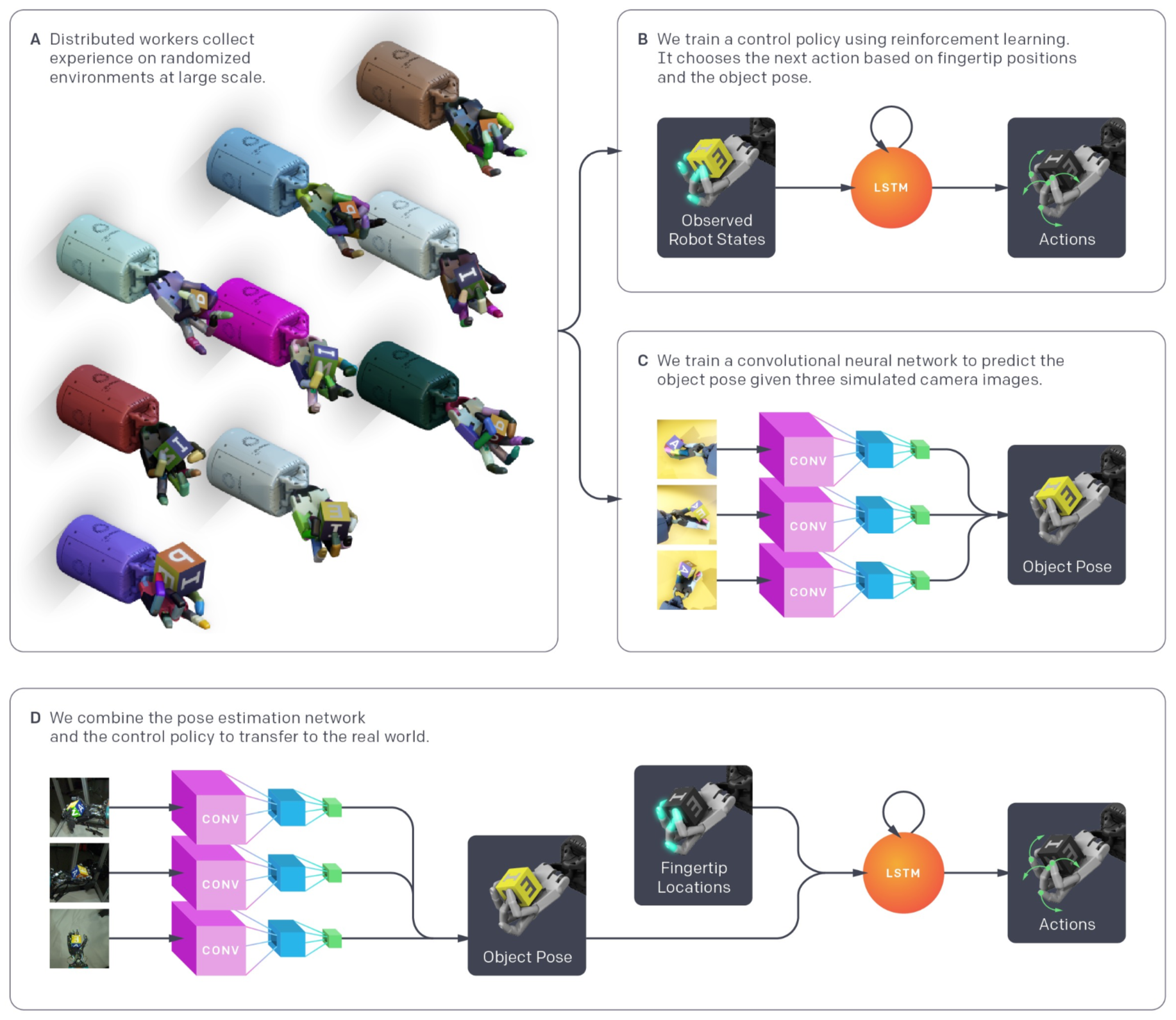}
\caption{Learning dexterous in-hand manipulation system overview. (from \cite{OpenAI2018})}
\label{DexterousRobot}
\end{figure}

There are four steps to learn dexterous in-hand manipulation as illustrated in Figure~\ref{DexterousRobot}:
1) collect experience;
2) train a policy using RL;
3) predict the object pose;
and 4) combine the pose estimation and the policy to deploy on the robot.

In Step 1, we collect data using a large number of workers with randomized parameters and appearances in simulation to learn the policy and to estimate vision-based poses.
Many aspects of the simulated environment are randomized, 
including observation, physics, unmodeled effects for imperfect actuation, and visual appearance.
In Step 2, we learn a policy based on the collected experience, using RL with a recurrent neural network (RNN),
in particular, the same distributed implementation of Proximal Policy Optimization (PPO) to train OpenAI Five for Dota 2,
where 128,000 CPUs and 256 GPUs were involved for training. 
In Step 3, we learn to predict the pose of the object from images with a convolutional neural network (CNN).
Policy and pose estimation are learned separately.
In Step 4, we deploy the policy and pose estimation learned with simulation to the real robot,
where robot fingertip locations are measured using a 3D motion capture system and
three real camera feeds are used to predict the object pose.
The policy can learn various grasp types, 
in particular, tip pinch grasp, palmar pinch grasp, tripod grasp, quadpod grasp, 5-finger precision grasp, and power grasp.

\subsection{Legged Robot}

Legged robots, in contrast to tracked or wheeled ones, 
are feasible in rough terrain and complex cluttered environments,
e.g., in forests, in mountains, on stairs, unstructured tunnels, and other planets.

The observations are the measurement of robot states, including height, linear velocity of the base, angular velocity of the base, positions and velocities of the joints, a sparsely sampled joint state history, and the previous action.
Actions are low-impedance joint position commands to the actuators.
For command-conditioned and high-speed locomotion, a command is represented by three components: forward velocity, lateral velocity, and the turning rate, the three desired body velocities.
Rewards are specified differently for different tasks.
%The readers are referred to the paper in Annotated Bibliography for more details. 

\begin{figure}[h]
\centering
\includegraphics[width=0.8\linewidth]{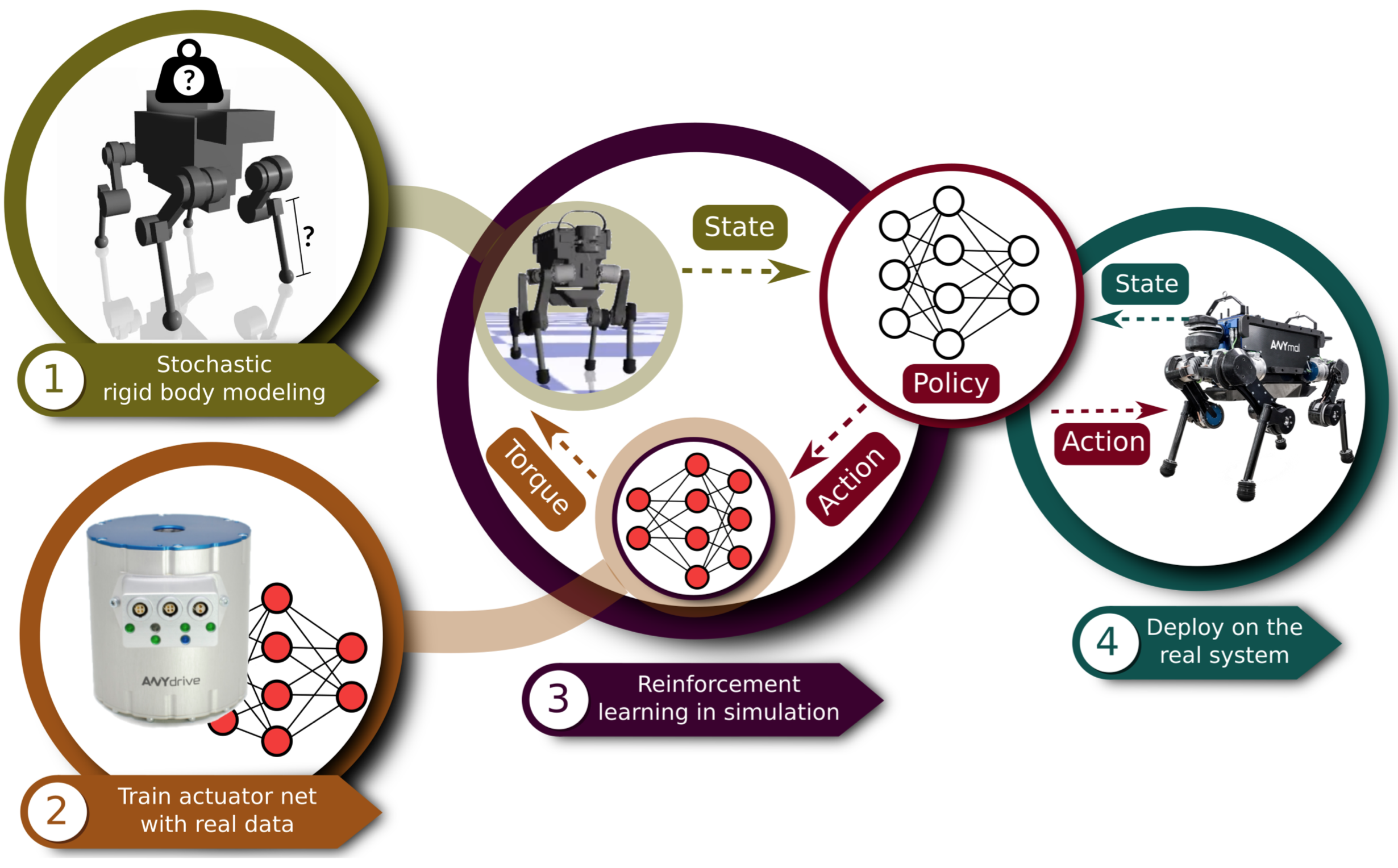}
\caption{Steps to create a policy for a legged robot. (from \cite{Hwangbo2019})}
\label{LeggedRobot}
\end{figure}

We study the ANYmal robot, a sophisticated quadrupedal system of medium dog size.
There are four steps to learn agile and dynamic motor skills for legged robots as illustrated in Figure~\ref{LeggedRobot}:
1) stochastic rigid body modelling;
2) train actuator net with real data; 
3) RL in simulation;
and 4) deploy on the real system.

In Step 1, well-known physical principles are used to model an idealized multibody system to resemble the rigid links of ANYmal.
In Step 2, an actuator net is learned end-to-end to map actuator actions to the resulting torques.
We can use a personal computer with one CPU and one GPU for training in a reasonably short time with decent performance.
The full hybrid simulator includes a rigid-body simulation and the actuator net.
In Step 3, we use the simulator and RL to train the controller, represented by a multilayer perceptron. 
In particular, Trust Region Policy Optimization (TRPO) is used for training.
Randomization is applied to some system properties, e.g., mass position center, link masses, joint positions, and kinematics. 
Then, in Step 4, we deploy the trained controller directly on the physical system by sim-to-real transfer.
The computation time for the inference of the simple network, i.e., for deciding on the next action, 
is at the level of 25 $\mu s$ on a single CPU thread,
which is much less than the computation time for model-based methods.

%Creating a control policy. In the first step, we identify the physical parameters of the robot and estimate uncertainties in the identification. In the second step, we train an actuator net that models complex actuator/software dynamics. In the In the fourth step, we deploy the trained policy directly on the physical system

\subsection{Annotated Bibliography}

The general discussion in the beginning is based on a survey of RL in robotics~\citep{Kober2013}.

The discussion of dextrous in-hand manipulation is based on \cite{OpenAI2018}.
See also \url{https://blog.openai.com/openai-five/}

Robot dexterity is among MIT Technology Review 10 Breakthrough Technologies 2019; 
see \url{https://www.technologyreview.com/lists/technologies/2019/}.
See also \url{https://bair.berkeley.edu/blog/2018/08/31/dexterous-manip/}

The discussion of legged robots is based on \cite{Hwangbo2019}.

Watch the NIPS 2017 Invited Talk on deep learning for robotics at \url{http://goo.gl/oVonGS}.
It is worthwhile to monitor researchers in robotics, e.g., Pieter Abbeel and Chelsea Finn among many others.

There are many efforts applying RL to robotics, including sim-to-real, imitation learning, value-based learning, policy-based learning, and model-based learning. See \cite{Li2017DeepRL} for more details.

\clearpage

\section{Transportation}
\label{transportation}

Transportation is related to everybody. 
RL is an approach to improve its efficiency and to reduce its cost. In the following, we discuss ridesharing order dispatching.

\subsection{Ridesharing Order Dispatching}

Online ride hailing services like Didi Chuxing and Uber have been changing the transportation dramatically, 
with a huge potential of transportation efficiency improvement.
Driver-passenger order dispatching is among the critical tasks like demand prediction, route planning, and fleet management of such ride sharing services.
It makes decisions to assign available drivers to unassigned passengers nearby,
considering spatial extent and temporal dynamics of the dispatching process.

Hierarchical RL is an approach to knowledge representation with temporal abstraction at multiple levels, and to learn and plan.
Options are a framework, where each option may span several raw actions.
Options are formulated as a semi-Markov decision process (SMDP).
A distributed state representation layer with cerebellar value networks can help improve stability of value iteration with nonlinear function approximation.
To combat with adversarial perturbation and noises, a regularized policy evaluation can help achieve robustness.
Transfer learning techniques can help adapt a learned policy to multiple cities. 
The learned policy show promising performance results w.r.t. metrics about total driver income and user experience on the platform of Didi Chuxing.

We next discuss the problem formulation for ride-sharing order dispatching.
A state is composed of a driver's geographical status, the raw time stamp, and the contextual feature vector.
Contextual features may be dynamic features for supply-demand around certain spatiotemporal point,
and static features like day of week, driver service statistics, and holiday indicator.
An option represents the transition of the driver from a spatiotemporal status to another in multiple time steps.
A transition with a trip assignment, a nonzero reward comes with the termination of the option.
A transition with an idle movement results with zero reward.
A policy specifies the probability of taking an option in a state.
A policy corresponds with a state value function.
And the goal of an RL algorithm is to estimate an optimal policy and/or its value function. 

\begin{figure}[h]
\centering
\includegraphics[width=0.8\linewidth]{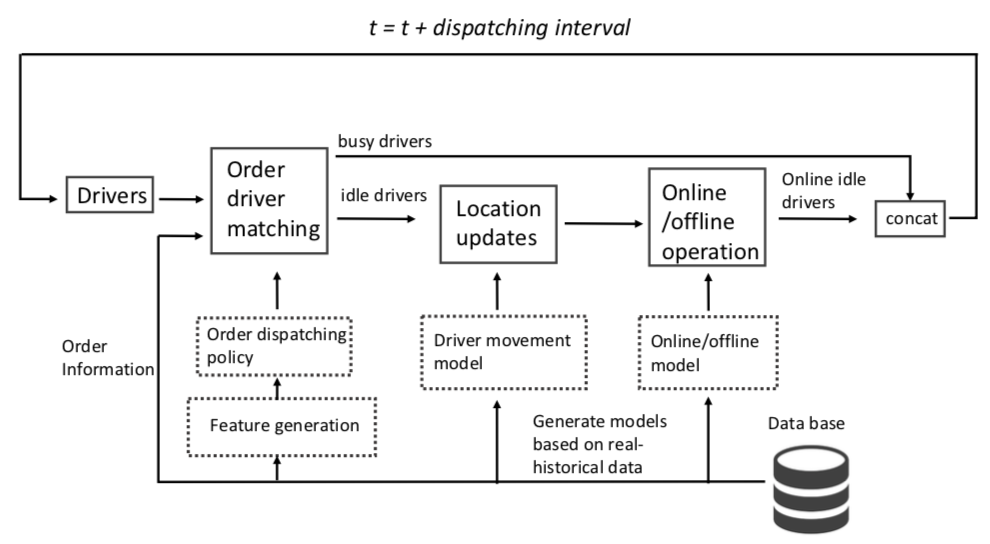}
\caption{Composition and workflow of the order dispatching simulator. (from ~\cite{Tang2019KDD})}
\label{OrderDispatching}
\end{figure}

A high-fidelity simulator is helpful for RL, e.g., generating synthetic data to evaluate or optimize a policy.
Figure~\ref{OrderDispatching} illustrates the composition and workflow of the order dispatching simulator built from historical real data.

At the beginning of the simulation, drivers' status and order information are initialized with historical real data.
After that, the simulator determines drivers' status, 
following an order-driver matching algorithm, 
with the help of an order dispatching policy learned with RL.
A busy driver will fulfill an order.
An idle driver follows a random walk, according to a driver movement model, 
and be in an online/offline mode, according to an online/offline model.
These two models are learned from the historical real data.

%\subsection{Traffic Light Control}

\subsection{Annotated Bibliography}

The discussion about multi-driver order dispatching is based on~\cite{Tang2019KDD}.
See also \cite{LiMinne2019WWW} for ridesharing order dispatching with an multi-agent RL approach.
See a tutorial about Deep Reinforcement Learning with Applications in Transportation at AAAI 2019 at \url{https://outreach.didichuxing.com/tutorial/AAAI2019/}. Check its updates at ICAPS 2019 and KDD 2019.

See \cite{Sutton1999} for discussions about hierarchical RL, options, and semi-MDP. 

See \cite{Simao2009} for an approximate dynamic programming for large-scale fleet management.
See \cite{WeiHua2018KDD} for an RL approach for intelligent traffic light control.

%Reinforcement Learning and Deep Learning based Lateral Control for Autonomous Driving \cite{LiDong2019}
%Autonomous reinforcement learning on raw visual input data in a real world application \cite{Lange2012}

\clearpage

%\input{biblio}              % bibliography
%{\footnotesize
%  \bibliography{DeepRL}
%}
%\bibliographystyle{apa} 

\addcontentsline{toc}{section}{Reference}
\bibliography{DeepRL}
\bibliographystyle{apa}

\end{document}